\pdfoutput=1
\documentclass[10pt,twocolumn,letterpaper]{article}

\usepackage[pagenumbers]{cvpr} %

\input{preamble}

\definecolor{cvprblue}{rgb}{0.21,0.49,0.74}
\usepackage[pagebackref,breaklinks,colorlinks,allcolors=cvprblue]{hyperref}
\definecolor{grapeurl}{HTML}{EC008C}
\hypersetup{urlcolor=grapeurl}
\let\GrapeOriginalTTDefault\ttdefault
\usepackage[varl,noupquote]{inconsolata}
\let\ttdefault\GrapeOriginalTTDefault
\newcommand{\GrapeCodeFont}{\fontencoding{T1}\fontfamily{zi4}\selectfont}
\renewcommand{\UrlFont}{\GrapeCodeFont}
\DeclareTextFontCommand{\texttt}{\GrapeCodeFont}
\AtBeginDocument{%
  \crefname{figure}{Figure}{Figures}%
  \Crefname{figure}{Figure}{Figures}%
}

\usepackage{graphicx}

\DeclareRobustCommand{\grapeicon}{%
  \raisebox{-1.5ex}{%
    \includegraphics[height=2.3em]{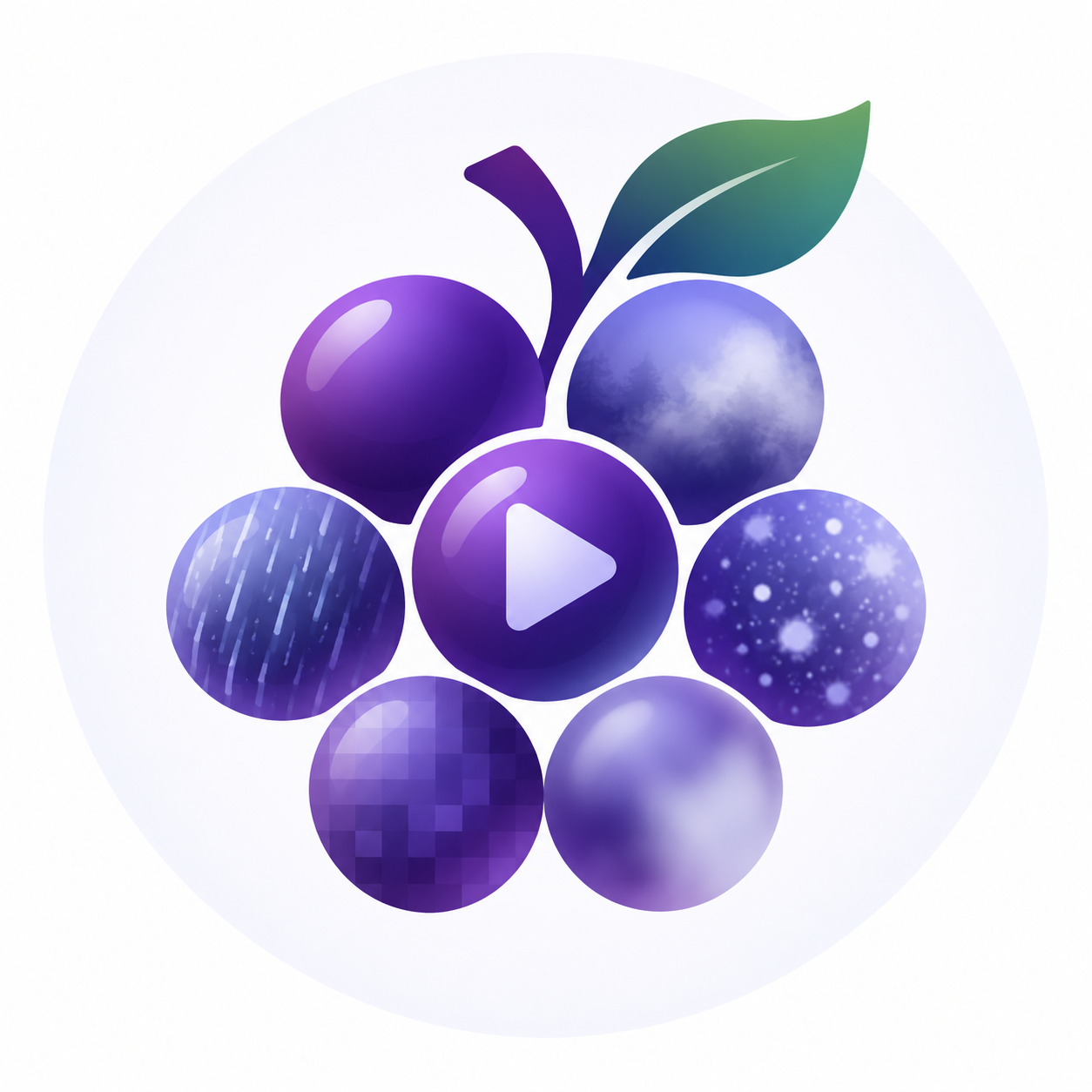}%
  }%
}

\usepackage{graphicx,booktabs,multirow,array}
\usepackage{xcolor,colortbl}
\usepackage{tikz}

\def\confName{CVPR}
\def\confYear{2027}

\title{\grapeicon\hspace{0.03em} \grapebench{}: When Vision Degrades, Does Video Reasoning Hold?}

\author{Ce Zhang$^{1*}$ \quad Junyi He$^{1*}$ \quad Tianshi Ming$^{1*}$ \quad Jinxi He$^{1}$ \quad Jianshu Zhang$^{2}$ \quad Xiyuan Yang$^{3}$ \\
Jianuo Zhu$^{1}$ \quad Yicheng Tao$^{1}$ \quad Xiyu Luo$^{1}$ \quad Minghao Fu$^{4}$ \quad Jiahui Zhu$^{5}$ \quad Kehan Guo$^{6}$ \\
Yunzhong Xiao$^{1}$ \quad Jingyang Lin$^{7}$ \quad Wei Chow$^{8}$ \quad Lingdong Kong$^{8}$ \quad Katia Sycara$^{1}$ \quad Yaqi Xie$^{3}$ \\[3pt]
$^{1}$CMU \quad $^{2}$Northwestern \quad $^{3}$UIUC \quad $^{4}$UCSD \quad $^{5}$WSU \quad $^{6}$ND \quad $^{7}$UR \quad $^{8}$NUS \quad $^{*}$Equal contribution \\[3pt]
Project page: \url{https://zhangce01.github.io/grape-bench/} \\
Code: \ghicon\ \href{https://github.com/zhangce01/GRAPE-Bench}{\UrlFont GRAPE-Bench} \quad Datasets: \hficon\ \href{https://huggingface.co/datasets/grape-lab/GRAPE-Bench-Synthetic}{\UrlFont GRAPE-Bench-Synthetic} \quad \hficon\ \href{https://huggingface.co/datasets/grape-lab/GRAPE-Bench-Realworld}{\UrlFont GRAPE-Bench-Realworld}
}

\begin{document}
\maketitle
\begin{abstract}
Despite strong performance on standard benchmarks, video reasoning models remain vulnerable to visual degradation, raising concerns about their reliability in safety-critical scenarios. Existing benchmarks address this concern in part by simulating visual degradation or probing model reasoning with challenging questions, leaving open how well models handle naturally degraded video and whether they can make the simpler judgments that matter for safe driving. To this end, we introduce \textbf{\grapebench{}}, which jointly evaluates video reasoning across visual degradation conditions and question difficulty. It combines a synthetic set spanning twelve corruption types across daily-life and driving videos with real-world footage of driving at night or in rain, snow, or fog. Each video is paired with three progressively challenging questions, from fundamental judgments with potential safety implications to demanding reasoning stress tests.
Across open-weight and frontier models, we observe ranking reversals between synthetic and real-world conditions and level-specific weaknesses obscured by overall accuracy. Synthetic-data fine-tuning improves Qwen3.5-9B's real-world accuracy from 76.7\% to 81.1\%, while clean-privileged on-policy distillation reaches 80.4\% in a separate reasoning-enabled setting. Through preference steering, we further demonstrate continuous control over a frozen model's trade-off between safety-relevant judgments and challenging reasoning, with little change in overall accuracy. Together, \grapebench{} provides a unified testbed for evaluating, improving, and controlling video reasoning under degraded visual conditions. 
\vspace{-15pt}
\end{abstract}

\begin{table*}[t]
\centering
\caption{\textbf{Comparison with existing benchmarks.} We compare \mbox{\benchdot[0.45em]{benchours}\grapebench{}} with \mbox{\benchdot[0.45em]{benchimage}image}, \mbox{\benchdot[0.45em]{benchvideo}general-video}, and \mbox{\benchdot[0.45em]{benchdriving}driving} benchmarks in scale (\mbox{\benchV\kern0.2em video samples}, \mbox{\benchI\kern0.2em images or synchronized image sets}), perturbation coverage, scenarios, and question design. \emph{Natural} denotes naturally degraded evaluation data; \emph{Difficulty Tiers} refers to question difficulty. Our video and QA counts list synthetic + real-world subsets.}
\vspace{-7pt}
\label{tab:benchmark-comparison}
\begingroup
\fontsize{7.75}{9.5}\selectfont
\setlength{\tabcolsep}{5.4pt}
\renewcommand{\arraystretch}{1}
\definecolor{benchgreen}{RGB}{0,140,0}
\definecolor{benchred}{RGB}{210,30,30}
\newcommand{\benchheadicon}[1]{%
  \begingroup
  \setbox0=\hbox{X}%
  \raisebox{0.5\ht0}[0pt][0pt]{%
    \begin{tikzpicture}[baseline=(current bounding box.center),
      x=1em,y=1em,draw=black!65,line width=0.45pt,
      line cap=round,line join=round]
      \path[use as bounding box] (0,0) rectangle (1,0.8);
      #1
    \end{tikzpicture}}%
  \endgroup\hspace{0.25em}}
\newcommand{\benchsamplesicon}{\benchheadicon{%
  \draw[rounded corners=0.45pt] (0.23,0.76) -- (0.96,0.76) -- (0.96,0.22);
  \draw[rounded corners=0.45pt] (0.04,0.04) rectangle (0.77,0.59);}}
\newcommand{\benchqaicon}{\benchheadicon{%
  \draw[rounded corners=0.45pt] (0.08,0.76) -- (0.96,0.76)
    -- (0.96,0.22) -- (0.41,0.22) -- (0.22,0.04)
    -- (0.22,0.22) -- (0.08,0.22) -- cycle;
  \node[inner sep=0pt,text=black!65,font=\fontsize{5.5}{6}\selectfont\bfseries]
    at (0.52,0.49) {?};}}
\newcommand{\bvid}[1]{#1\kern0.28em\benchV}
\newcommand{\bimg}[1]{#1\kern0.28em\benchI}
\newcommand{\byes}{\textcolor{benchgreen}{\ding{51}}}
\newcommand{\bno}{\textcolor{benchred}{\ding{55}}}
\begin{tabular*}{\textwidth}{@{\extracolsep{\fill}}l>{\upshape}ccccccccc@{}}
\toprule
\multirow{2}{*}{\textbf{Benchmark}}
& \multirow{2}{*}{\textbf{Venue}}
& \multicolumn{2}{c}{\textbf{Scale}}
& \multicolumn{3}{c}{\textbf{Perturbation Properties}}
& \multicolumn{2}{c}{\textbf{Scenarios}}
& \multicolumn{1}{c}{\textbf{Question Design}} \\
\cmidrule(lr){3-4}\cmidrule(lr){5-7}\cmidrule(lr){8-9}\cmidrule(l){10-10}
& & \#Visual Data & \#QAs & \#Types & Synthetic & Natural & Daily Life & Driving & Difficulty Tiers \\
\midrule
\benchdot{benchimage}ImageNet-C~\cite{hendrycks2019imagenetc} & \textit{ICLR\,'19} & \bimg{50,000} & -- & 15 & \byes & \bno & \byes & \bno & \bno \\
\benchdot{benchvideo}Video-MME~\cite{fu2025videomme} & \textit{CVPR\,'25} & \bvid{900} & 2,700 & 0 & \bno & \bno & \byes & \bno & \bno \\
\benchdot{benchvideo}Video-MME-v2~\cite{fu2026videommev2} & \textit{arXiv\,'26} & \bvid{800} & 3,200 & 0 & \bno & \bno & \byes & \bno & \byes \\
\benchdot{benchvideo}MVTamperBench~\cite{agarwal2025mvtamperbench} & \textit{ACL-F\,'25} & \bvid{17,435} & 19,000 & 5 & \byes & \bno & \byes & \bno & \bno \\
\benchdot{benchvideo}PVRBench~\cite{he2026rova} & \textit{ECCV\,'26} & \bvid{$\sim$9,000} & $\sim$52,000 & 12 & \byes & \bno & \byes & \byes & \bno \\
\benchdot{benchdriving}DriveLM-nuScenes~\cite{sima2024drivelm}
& \textit{ECCV\,'24} & \bimg{4,871} & 393,436
& 0 & \bno & \bno & \bno & \byes & \bno \\
\benchdot{benchdriving}DriveBench~\cite{xie2025drivebench} & \textit{ICCV\,'25} & \bimg{3,200} & 24,772 & 15 & \byes & \bno & \bno & \byes & \bno \\
\benchdot{benchdriving}ObsDriveBench~\cite{yan2026obsdrivebench} & \textit{arXiv\,'26} & \bimg{2,568} & 28,136 & 4 & \bno & \byes & \bno & \byes & \bno \\
\midrule
\benchdot{benchours}\textbf{\grapebench{} (Ours)} & -- & \bvid{\textbf{10,841 + 572}} & \textbf{32,523 + 1,716} & \textbf{8 + 4} & \byes & \byes & \byes & \byes & \byes \\
\bottomrule
\end{tabular*}
\endgroup
\vspace{-8pt}
\end{table*}

\section{Introduction}
\label{sec:intro}

\begin{figure}[t]
    \centering
    \includegraphics[width=\columnwidth]{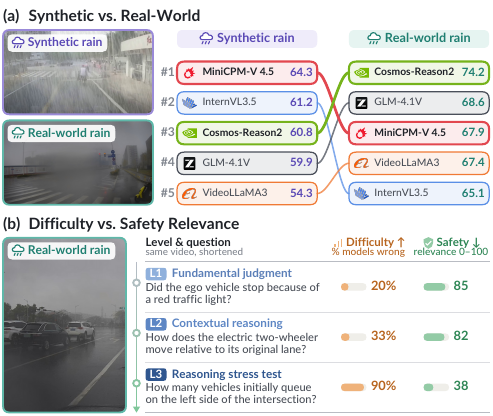}
    \vspace{-15pt}
    \caption{\textbf{Two observations motivating \grapebench{}.}
(a) Model rankings change between driving videos with synthetic rain and footage captured in real rain. Among the five models shown, MiniCPM-V 4.5 drops from first to third, while Cosmos-Reason2 rises from third to first.
(b) Three questions about the same rainy driving scene illustrate why greater difficulty need not imply greater safety relevance. Identifying whether the ego vehicle stopped for a red light (L1) and reasoning about a two-wheeler's motion relative to its original lane (L2) both receive high safety relevance scores. Counting the vehicles initially queued on the left (L3) is the hardest question, answered incorrectly by 90\% of the evaluated models, yet receives a safety relevance score of 38.}
    \label{fig:teaser}
    \vspace{-10pt}
\end{figure}

Recent advances in multimodal large language models (MLLMs) have improved temporal perception and reasoning on video understanding benchmarks such as MVBench~\cite{li2024mvbench} and Video-MME~\cite{fu2025videomme}. In real-world environments, however, adverse weather, low illumination, motion blur, and sensor disturbances can obscure the visual cues on which these judgments depend~\cite{sakaridis2021acdc,burnett2023boreas}. Prior studies have documented failures in both visual perception and multimodal reasoning under such degradations~\cite{dong2023corrupt3d,qiu2025mllmic,saxena2026vlmrobustbench,he2026rova}. In driving evaluations, models can give plausible answers with little grounding in the visual evidence, raising safety concerns when such answers inform driving decisions~\cite{xie2025drivebench}. For example, a model that overlooks a pedestrian stepping off the curb in heavy rain may report that the crosswalk is clear when the approaching driver needs to yield.

Existing video robustness benchmarks use controlled spatiotemporal corruptions to evaluate how models respond as visual conditions deteriorate~\cite{agarwal2025mvtamperbench,he2026rova,xie2025drivebench}. Such tests allow systematic comparisons, but the resulting model rankings need not carry over to naturally degraded footage. In Figure~\ref{fig:teaser}(a), for example, MiniCPM-V 4.5 outperforms Cosmos-Reason2 on videos with artificially added rain, while their ordering reverses on footage recorded in real rain. Within a given video, the choice of questions shapes which capabilities are tested. Video understanding benchmarks use challenging questions to probe how well MLLMs integrate evidence over time~\cite{li2024mvbench,fu2025videomme}. In driving, however, the questions that models find hardest need not be the most relevant to a driving decision. In Figure~\ref{fig:teaser}(b), models find it harder to count the vehicles initially queued on the left side of the intersection than to determine whether the ego vehicle stopped for a red light. Of the two, the traffic signal bears more directly on the decision to stop. This motivates pairing questions at different levels of reasoning demand within the same video, so that an evaluation can probe the limits of a model's reasoning while still testing the fundamental judgments that matter for safe driving.

In this work, we introduce \textbf{\grapebench{}} (\textbf{G}raded \textbf{R}obustness \textbf{A}ssessment under \textbf{P}erturbed \textbf{E}nvironments), a tiered benchmark for evaluating video understanding under visual degradation (Table~\ref{tab:benchmark-comparison}). The synthetic track applies a total of twelve corruption types across daily-life and driving videos, while the real-world track uses driving footage recorded in rain, snow, or fog, as well as at night. Across both tracks, each video is paired with three independently answerable questions designed to increase in reasoning demand, ranging from fundamental, potentially safety-critical judgments to demanding reasoning stress tests. As summarized in Table~\ref{tab:benchmark-comparison}, \grapebench{} jointly covers synthetic and natural degradations, diverse scenarios, and difficulty-graded questions, enabling systematic evaluation across visual conditions and reasoning demands. 

We evaluate 23 open-weight models on both the synthetic and real-world sets, with additional results for proprietary frontier models on the real-world set. In driving scenes, some models outperform their peers under a simulated adverse condition but fall behind those same peers on footage recorded under the corresponding natural condition. We also find that some models achieve higher average accuracy across the three tiers despite lower accuracy on questions that test fundamental scene understanding. Their advantage on harder questions can therefore mask weaknesses in basic judgments that matter for safe driving, such as recognizing when a vehicle needs to stop or yield.
\looseness=-1 %

Beyond evaluation, the synthetic set pairs each corrupted video with its clean counterpart, with the same questions for both versions. We use these data to study whether training on synthetic corruptions improves performance under natural degradation. Supervised fine-tuning (SFT) on the synthetic set raises Qwen3.5-9B's accuracy on the real-world set from 76.7\% to 81.1\%. In the reasoning-enabled setting, on-policy self-distillation (OPSD) with privileged access to the clean video during training raises real-world accuracy from 78.1\% to 80.4\%. The question tiers also let us test whether a model can be steered toward fundamental, safety-relevant judgments or toward more demanding reasoning tasks. With the underlying VLM frozen, we learn a preference-conditioned steering mechanism. On held-out driving videos from both sets, varying the steering coefficient shifts the balance of accuracy between these two types of questions, with little change in overall accuracy.
\looseness=-1

\begingroup
\setlength{\parskip}{0pt}
Our contributions can be summarized as follows:
\begin{itemize}[topsep=3pt,itemsep=3pt,partopsep=0pt]
  \item We introduce \grapebench{}, combining synthetic corruptions and naturally adverse driving conditions with a three-level question hierarchy spanning safety-critical perception and demanding reasoning.
  \item We evaluate open-weight and frontier proprietary models, showing that synthetic advantages can disappear under natural conditions and that aggregate accuracy can conceal weaknesses at individual question levels.
  \item We demonstrate that SFT and OPSD on synthetic data improve real-world accuracy, and that lightweight steering vectors control the balance between safety-critical perception and demanding reasoning in a frozen model.
\end{itemize}
\endgroup

\vspace{-3.5pt}
\section{Motivation}
\label{sec:motivation}

\begin{figure}[t]
    \centering
    \includegraphics[width=\columnwidth]{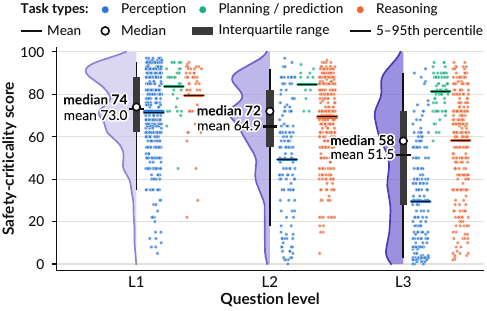}
    \vspace{-15pt}
    \caption{\textbf{Mean safety criticality decreases with difficulty level.} Safety-criticality scores (0--100) for 1,716 real-world questions (572 per level), rated by \claudeicon\nobreak\hspace{0.2em}Claude Opus~5.5. Half violins show score distributions; black ticks and open circles mark means and medians, while thick bars and whiskers indicate the interquartile and 5th--95th percentile ranges. Colored dots and short lines denote individual questions and means by task type.}
    \label{fig:safety_levels}
    \vspace{-12pt}
\end{figure}

\noindent\textbf{Beyond Synthetic Corruptions.}
Synthetic corruptions provide a controlled setting for evaluating visual robustness by perturbing video inputs while preserving their underlying content and questions. Existing benchmarks, including MVTamperBench, PVRBench, and DriveBench, have adopted such perturbations to probe model reliability~\cite{agarwal2025mvtamperbench,he2026rova,xie2025drivebench} (Table~\ref{tab:benchmark-comparison}). However, relative model performance under synthetic perturbations need not persist under naturally adverse conditions. We observe a ranking reversal in \grapebench{}: MiniCPM-V~4.5-8B outperforms Cosmos-Reason2-8B across all four synthetic driving corruptions, yet their ranking reverses under each corresponding real-world condition (Tables~\ref{tab:grape_main_levels} and~\ref{tab:grape_realworld_levels}). Figure~\ref{fig:teaser}(a) further illustrates this discrepancy by juxtaposing representative synthetic and real-world rain footage with model rankings based on aggregate accuracy over the corresponding evaluation subsets. These rank reversals motivate evaluating models under both controlled corruptions and naturally adverse driving conditions.

\vspace{3pt} \noindent
\textbf{Difficulty vs. Safety Relevance.}
Beyond the conditions under which models fail, an equally important question is which judgments are affected. Reasoning difficulty does not necessarily reflect safety relevance: identifying an immediate road hazard may require only basic visual perception, whereas a more demanding reasoning task may concern incidental scene details. This distinction motivates our three-level question hierarchy, progressing from fundamental, potentially safety-critical judgments to challenging reasoning stress tests. Figure~\ref{fig:teaser}(b) illustrates this distinction through three questions from the same driving video, where increasing reasoning difficulty coincides with decreasing assessed safety relevance. To examine this pattern more broadly, we analyze the 1,716 questions we construct for the real-world driving set of \grapebench{} (Section~\ref{sec:data-construction}). We use Claude Opus~5.5~\cite{anthropic2026claudeopus55} to assign each question an auxiliary safety-relevance score. As shown in Figure~\ref{fig:safety_levels}, the average ratings decrease from 73.0 at L1 to 64.9 at L2 and 51.5 at L3, with L1 rated higher than L3 in 408 of 572 video groups (71.3\%). These observations illustrate why reasoning difficulty alone cannot determine which judgments deserve attention. By evaluating accuracy separately at each level, \grapebench{} reveals weaknesses in fundamental, potentially consequential judgments that may otherwise be obscured by overall performance.

\section{\grapebench{}}
\label{sec:grapebench}

\begingroup
\definecolor{GrapeDriving}{HTML}{7460C2}
\definecolor{GrapeDaily}{HTML}{279B8A}
\definecolor{GrapeAvg}{HTML}{D99C39}
\definecolor{GrapeDelta}{HTML}{C43E3E}
\definecolor{GrapeText}{HTML}{20232A}
\definecolor{GrapeMuted}{HTML}{70757E}
\definecolor{GrapeRule}{HTML}{C6C9D0}
\definecolor{GrapeBand}{HTML}{EEEEEE}
\newcommand{\GrapeFirst}[2]{\cellcolor{#1!38}\textbf{#2}}
\newcommand{\GrapeSecond}[2]{\cellcolor{#1!24}\underline{#2}}
\newcommand{\GrapeThird}[2]{\cellcolor{#1!12}#2}
\newcommand{\GrapeDeltaFirst}[1]{\GrapeFirst{GrapeDelta}{\textcolor{GrapeDelta}{#1}}}
\newcommand{\GrapeDeltaSecond}[1]{\textcolor{GrapeDelta}{\GrapeSecond{GrapeDelta}{#1}}}
\newcommand{\GrapeDeltaThird}[1]{\GrapeThird{GrapeDelta}{\textcolor{GrapeDelta}{#1}}}
\DeclareRobustCommand{\GrapeKey}[2]{%
  \begingroup\setlength{\fboxsep}{0pt}%
  \raisebox{0pt}[\ht\strutbox][\dp\strutbox]{%
    \colorbox{GrapeDriving!#1}{\hspace{2pt}\rule[-0.3ex]{0pt}{2ex}#2\hspace{2pt}}}%
  \endgroup}

\makeatletter
\newcommand{\GrapeRowCenter}[1]{\raisebox{\dimexpr(\ht\@arstrutbox-\dp\@arstrutbox-\height+\depth)/2\relax}[0pt][0pt]{#1}}
\makeatother
\newsavebox{\GrapeModelNameBox}
\newlength{\GrapeModelAxis}
\newcommand{\GrapeBadgeWidth}{11pt}
\newcommand{\GrapeBadgeHeight}{9.8pt}
\newcommand{\GrapeModelIndent}{14pt}
\newlength{\GrapeBadgeOpticalShift}
\newcommand{\GrapeBadgeShiftkimi}{\dimexpr\GrapeBadgeWidth*3/25\relax}
\newcommand{\GrapeBadge}[1]{%
  \begingroup
  \setlength{\GrapeBadgeOpticalShift}{0pt}%
  \ifcsname GrapeBadgeShift#1\endcsname
    \setlength{\GrapeBadgeOpticalShift}{\csname GrapeBadgeShift#1\endcsname}%
  \fi
  \raisebox{\dimexpr\GrapeModelAxis-(\height-\depth)/2\relax}[0pt][0pt]{%
    \makebox[\GrapeBadgeWidth][c]{\hspace*{\GrapeBadgeOpticalShift}%
      \includegraphics[width=\GrapeBadgeWidth,height=\GrapeBadgeHeight,keepaspectratio]{logos/#1.pdf}%
      \hspace*{-\GrapeBadgeOpticalShift}}}%
  \endgroup}

\newcommand{\GrapeIcon}[2]{\begin{tikzpicture}[x=10pt,y=10pt,draw=#1,fill=#1,line width=.58pt,line cap=round,line join=round]
\path[use as bounding box] (-.04,-.06) rectangle (1.04,1.02);
\csname GrapeIconDraw#2\endcsname
\end{tikzpicture}}
\newcommand{\GrapeCloud}{\draw[fill=white] (.19,.45)..controls(-.05,.45)and(-.02,.78)..(.22,.77)..controls(.19,1.03)and(.65,1.10)..(.71,.81)..controls(1.03,.87)and(1.10,.46)..(.85,.45)--cycle;}
\newcommand{\GrapeIconDrawRain}{\GrapeCloud\draw (.26,.30)--(.18,.12) (.51,.30)--(.43,.12) (.76,.30)--(.68,.12);}
\newcommand{\GrapeIconDrawSnow}{%
\foreach \a in {0,60,...,300}{\draw[rotate around={\a:(.5,.5)}] (.5,.5)--(.5,.97) (.407,.883)--(.5,.79)--(.593,.883);}}
\newcommand{\GrapeIconDrawNight}{
\path[draw,fill opacity=.13] (.72,.95)..controls(.06,1.09)and(-.12,.09)..(.55,.08)..controls(.81,.06)and(.99,.25)..(.99,.43)..controls(.48,.2)and(.25,.78)..(.72,.95)--cycle;
\draw (.83,.8)--(.99,.8) (.91,.72)--(.91,.88);}
\newcommand{\GrapeIconDrawFog}{\draw (.12,.81)--(.86,.81) (.01,.58)--(.65,.58) (.28,.35)--(1,.35) (.07,.12)--(.76,.12);}
\newcommand{\GrapeIconDrawWet}{
\draw[rounded corners=.8pt] (.05,.09) rectangle (.94,.9);
\path[draw,fill opacity=.15] (.39,.79)..controls(.32,.65)and(.21,.53)..(.22,.43)..controls(.23,.22)and(.55,.22)..(.56,.43)..controls(.57,.53)and(.46,.65)..(.39,.79)--cycle;
\path[draw,fill opacity=.15] (.76,.67)..controls(.69,.53)and(.63,.45)..(.65,.38)..controls(.67,.26)and(.85,.28)..(.86,.38)..controls(.87,.45)and(.81,.56)..(.76,.67)--cycle;}
\newcommand{\GrapeIconDrawGlass}{
\draw[rounded corners=.8pt] (.08,.08) rectangle (.93,.94);
\draw (.23,.23)--(.77,.77) (.23,.53)--(.47,.77) (.54,.23)--(.77,.46);}
\newcommand{\GrapeIconDrawOverexp}{
\draw[fill opacity=.16] (.5,.5) circle (.23);
\foreach \a in {0,45,...,315}{\draw[rotate around={\a:(.5,.5)}] (.5,.86)--(.5,1.02);}}
\newcommand{\GrapeIconDrawDefocus}{
\draw[opacity=.35,line width=1.5pt] (.5,.5) circle (.41);
\draw[opacity=.6,line width=.9pt] (.5,.5) circle (.29);
\draw (.5,.5) circle (.16);}
\newcommand{\GrapeIconDrawMotion}{
\draw (.02,.78)--(.47,.78) (.09,.51)--(.37,.51) (.02,.24)--(.47,.24);
\draw[fill opacity=.12] (.7,.5) circle (.27);}
\newcommand{\GrapeIconDrawBitrate}{
\draw[rounded corners=.7pt] (.02,.1) rectangle (.98,.91);
\fill[opacity=.65] (.11,.60) rectangle (.34,.82);
\fill[opacity=.2] (.36,.60) rectangle (.60,.82);
\fill[opacity=.35] (.62,.35) rectangle (.88,.58);
\fill[opacity=.5] (.36,.17) rectangle (.60,.38);
\draw (.11,.34)--(.32,.34) (.62,.77)--(.88,.77);}

\newcommand{\GrapeIconDrawUnderexp}{
\draw[fill opacity=.55] (.5,.5) circle (.23);
\foreach \a in {0,45,...,315}{\draw[rotate around={\a:(.5,.5)},opacity=.35] (.5,.86)--(.5,.95);}}
\newcommand{\GrapeIconDrawBiterror}{
\draw[rounded corners=.7pt] (.02,.1) rectangle (.98,.91);
\draw (.12,.52)--(.3,.52)--(.4,.3)--(.55,.74)--(.65,.52)--(.88,.52);
\fill[opacity=.45] (.14,.66) rectangle (.3,.8);
\fill[opacity=.3] (.68,.2) rectangle (.86,.36);}

\newcommand{\GrapeHead}[3]{\shortstack[c]{\GrapeIcon{#1}{#2}\\[-1pt]{\fontsize{8}{9}\selectfont #3}}}
\newcommand{\GrapeModel}[3]{%
  \begingroup
  \sbox{\GrapeModelNameBox}{\ifnum#2=1\textcolor{GrapeMuted}{#3}\else#3\fi}%
  \setlength{\GrapeModelAxis}{\dimexpr(\ht\GrapeModelNameBox-\dp\GrapeModelNameBox)/2\relax}%
  \makebox[\GrapeModelIndent][l]{#1}\usebox{\GrapeModelNameBox}%
  \endgroup}
\newcommand{\GrapeGroup}[1]{\rowcolor{GrapeBand}\multicolumn{13}{l}{\color{GrapeText}\bfseries #1} \\}
\newcommand{\GrapeCaption}{\textbf{\grapebench{} results under visual corruptions.} Accuracy (\%) with L1--L3 questions pooled. Rain, snow, night, and fog apply to driving videos (VRU-Accident); the other eight corruptions apply to daily-life videos. Avg is the overall question-level accuracy. Higher is better.}
\ifdefined\GrapeMeasureBox\else\newsavebox{\GrapeMeasureBox}\fi
\newcommand{\GrapeSection}[2]{\rowcolor{GrapeBand}\multicolumn{#1}{l}{\GrapeRowCenter{\color{GrapeText}\bfseries #2}} \\}
\newcommand{\GrapeCaptionMain}{\textbf{Results on the \grapebench{} synthetic set.} Accuracy (\%; higher is better) under twelve visual corruptions. Driving reports L1--L3 separately; daily-life columns (Avg.) pool questions across levels, and All pools questions across all levels and corruptions.\textcolor{GrapeDelta}{$\Delta$} is synthetic minus clean All accuracy (pp) on the same questions. Shading indicates \GrapeKey{38}{\textbf{1st}}, \GrapeKey{24}{\underline{2nd}}, and \GrapeKey{12}{3rd} place per column.}
\newcommand{\GrapeCaptionWide}{\textbf{\grapebench{} results by corruption and question level.} Accuracy (\%) for L1, L2, and L3 under each corruption. Rain, snow, night, and fog apply to driving videos; the other corruptions apply to daily-life videos. All corruptions pools every question of the corresponding level (10,841 per level). Higher is better.}
\newcommand{\GrapeCaptionStacked}{\textbf{\grapebench{} results by corruption and question level.} Each model occupies three rows, ordered L1, L2, and L3. All corruptions pools every question of the corresponding level. Rankings compare models within the same corruption and level.}

\begin{table*}[t]
\centering
\begin{minipage}{\textwidth}
\caption{\GrapeCaptionMain{}}
\vspace{-5pt}
\label{tab:grape_main_levels}
\begingroup
\newlength{\GrapeColW}\setlength{\GrapeColW}{19pt}
\renewcommand{\GrapeHead}[3]{\makebox[\GrapeColW][c]{\shortstack[c]{\GrapeIcon{#1}{#2}\\[-1pt]\parbox[c][14pt][c]{22pt}{\centering\fontsize{6}{6.6}\selectfont #3}}}}
\newcommand{\GrapeDailyHead}[3]{\GrapeHead{GrapeDaily}{#1}{#2\\#3}}
\renewcommand{\GrapeBadgeWidth}{8.5pt}
\renewcommand{\GrapeBadgeHeight}{7.9pt}
\renewcommand{\GrapeModelIndent}{11pt}
\begingroup
\setlength{\tabcolsep}{1.5pt}
\renewcommand{\arraystretch}{1.05}
\fontsize{7}{8.4}\selectfont
\color{GrapeText}
\sbox{\GrapeMeasureBox}{%
\begin{tabular}{@{}l@{\hspace{3pt}}c *{22}{>{\centering\arraybackslash}p{\GrapeColW}}}
\arrayrulecolor{GrapeText}\toprule[.8pt]
\multicolumn{1}{l}{\multirow{3}{*}[-15pt]{\textbf{Model}}} & \multirow{3}{*}[-15pt]{\textbf{Size}} & \multicolumn{12}{c}{\textcolor{GrapeDriving}{\textbf{Driving}}} & \multicolumn{8}{c}{\textcolor{GrapeDaily}{\textbf{Daily-life}}} & \multicolumn{1}{c}{\multirow{3}{*}[-15pt]{\textcolor{GrapeAvg!75!black}{\textbf{All}}}} & \multicolumn{1}{c}{\multirow{3}{*}[-15pt]{\textcolor{GrapeDelta}{$\boldsymbol{\Delta}$}}} \\
\arrayrulecolor{GrapeRule}\cmidrule(lr){3-14}\cmidrule(lr){15-22}
 & & \multicolumn{3}{c}{\GrapeHead{GrapeDriving}{Rain}{Rain}} & \multicolumn{3}{c}{\GrapeHead{GrapeDriving}{Snow}{Snow}} & \multicolumn{3}{c}{\GrapeHead{GrapeDriving}{Night}{Night}} & \multicolumn{3}{c}{\GrapeHead{GrapeDriving}{Fog}{Fog}} & \GrapeDailyHead{Wet}{Wet}{lens} & \GrapeDailyHead{Glass}{Glass}{occl.} & \GrapeDailyHead{Overexp}{Over}{exp.} & \GrapeDailyHead{Underexp}{Under}{exp.} & \GrapeHead{GrapeDaily}{Defocus}{Defocus} & \GrapeDailyHead{Motion}{Motion}{blur} & \GrapeDailyHead{Bitrate}{Low}{bitrate} & \GrapeDailyHead{Biterror}{Bit}{error} & & \\
\cmidrule(lr){3-5}\cmidrule(lr){6-8}\cmidrule(lr){9-11}\cmidrule(lr){12-14}\cmidrule(lr){15-22}
 & & L1 & L2 & L3 & L1 & L2 & L3 & L1 & L2 & L3 & L1 & L2 & L3 & \textcolor{GrapeDaily}{Avg.} & \textcolor{GrapeDaily}{Avg.} & \textcolor{GrapeDaily}{Avg.} & \textcolor{GrapeDaily}{Avg.} & \textcolor{GrapeDaily}{Avg.} & \textcolor{GrapeDaily}{Avg.} & \textcolor{GrapeDaily}{Avg.} & \textcolor{GrapeDaily}{Avg.} &  & \\
\arrayrulecolor{GrapeText}\midrule[.5pt]
\GrapeSection{24}{Open-Source VLMs}
\GrapeModel{\GrapeBadge{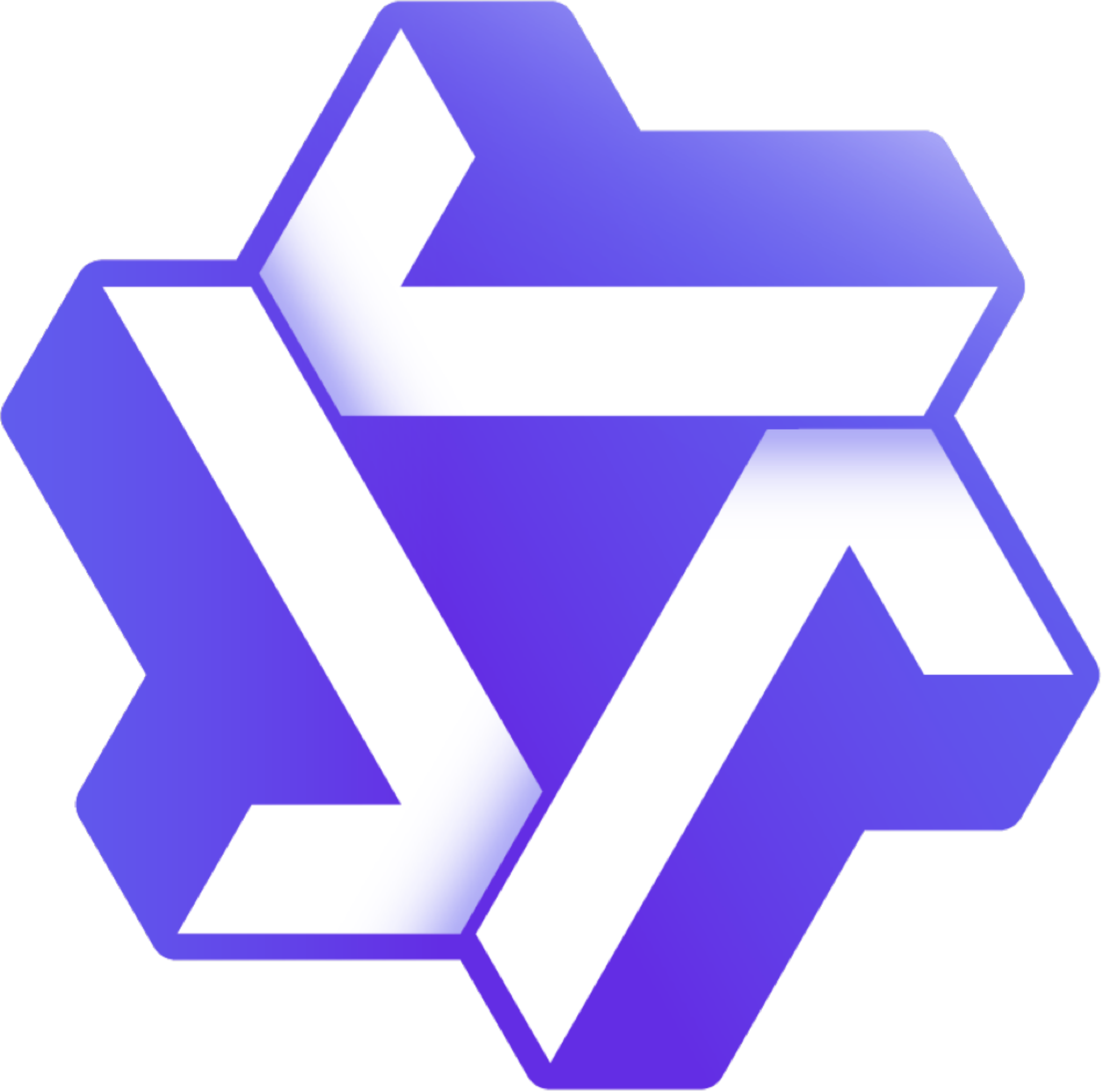}}{0}{Qwen3.8-Flash-Next} & 125B & \GrapeFirst{GrapeDriving}{82.9} & \GrapeThird{GrapeDriving}{74.9} & \GrapeThird{GrapeDriving}{66.1} & \GrapeFirst{GrapeDriving}{78.3} & 70.3 & \GrapeThird{GrapeDriving}{63.2} & \GrapeFirst{GrapeDriving}{84.3} & \GrapeThird{GrapeDriving}{74.8} & \GrapeThird{GrapeDriving}{70.7} & \GrapeFirst{GrapeDriving}{84.8} & 73.2 & \GrapeThird{GrapeDriving}{68.1} & \GrapeThird{GrapeDaily}{81.2} & 68.9 & \GrapeThird{GrapeDaily}{84.9} & \GrapeThird{GrapeDaily}{81.1} & \GrapeThird{GrapeDaily}{78.2} & \GrapeThird{GrapeDaily}{83.1} & \GrapeThird{GrapeDaily}{80.1} & \GrapeThird{GrapeDaily}{84.4} & \GrapeThird{GrapeAvg}{78.6} & \textcolor{GrapeDelta}{-9.0} \\
\GrapeModel{\GrapeBadge{qwen}}{0}{Qwen3.8} & 27B & 78.5 & 68.3 & 64.7 & 75.9 & 70.8 & 63.1 & 80.7 & 70.3 & 61.3 & 78.0 & 70.3 & 64.3 & 77.9 & \GrapeThird{GrapeDaily}{70.0} & 82.8 & 78.5 & 75.4 & 80.8 & 76.9 & 82.3 & 76.0 & \textcolor{GrapeDelta}{-9.1} \\
\GrapeModel{\GrapeBadge{qwen}}{0}{Qwen3.6} & 27B & \GrapeSecond{GrapeDriving}{82.1} & \GrapeFirst{GrapeDriving}{76.9} & \GrapeSecond{GrapeDriving}{68.0} & \GrapeSecond{GrapeDriving}{77.9} & \GrapeFirst{GrapeDriving}{76.7} & \GrapeSecond{GrapeDriving}{67.6} & \GrapeThird{GrapeDriving}{83.9} & \GrapeSecond{GrapeDriving}{78.5} & \GrapeSecond{GrapeDriving}{73.5} & 80.8 & \GrapeSecond{GrapeDriving}{77.2} & \GrapeFirst{GrapeDriving}{73.5} & \GrapeFirst{GrapeDaily}{82.4} & \GrapeFirst{GrapeDaily}{74.2} & \GrapeSecond{GrapeDaily}{85.4} & \GrapeFirst{GrapeDaily}{82.7} & \GrapeSecond{GrapeDaily}{79.1} & \GrapeSecond{GrapeDaily}{84.1} & \GrapeSecond{GrapeDaily}{80.5} & \GrapeFirst{GrapeDaily}{86.0} & \GrapeFirst{GrapeAvg}{80.3} & \GrapeDeltaFirst{-7.8} \\
\GrapeModel{\GrapeBadge{qwen}}{0}{Qwen3.5} & 27B & \GrapeSecond{GrapeDriving}{82.1} & \GrapeSecond{GrapeDriving}{75.6} & \GrapeFirst{GrapeDriving}{70.3} & \GrapeSecond{GrapeDriving}{77.9} & \GrapeSecond{GrapeDriving}{76.3} & \GrapeFirst{GrapeDriving}{67.7} & \GrapeSecond{GrapeDriving}{84.1} & \GrapeFirst{GrapeDriving}{78.9} & \GrapeFirst{GrapeDriving}{74.3} & \GrapeThird{GrapeDriving}{81.6} & \GrapeFirst{GrapeDriving}{77.7} & \GrapeSecond{GrapeDriving}{72.1} & \GrapeSecond{GrapeDaily}{82.0} & \GrapeSecond{GrapeDaily}{73.3} & \GrapeFirst{GrapeDaily}{85.8} & \GrapeSecond{GrapeDaily}{82.1} & \GrapeFirst{GrapeDaily}{80.0} & \GrapeFirst{GrapeDaily}{84.4} & \GrapeFirst{GrapeDaily}{80.8} & \GrapeFirst{GrapeDaily}{86.0} & \GrapeFirst{GrapeAvg}{80.3} & \GrapeDeltaSecond{-7.9} \\
\GrapeModel{}{1}{Qwen3.5} & 9B & 77.1 & 68.3 & 62.4 & 74.0 & 68.8 & 60.5 & 81.6 & 70.0 & 64.7 & 77.5 & 70.1 & 62.3 & 77.2 & 66.5 & 81.9 & 76.4 & 72.7 & 80.1 & 75.8 & 81.2 & 74.6 & \textcolor{GrapeDelta}{-9.4} \\
\GrapeModel{}{1}{Qwen3.5} & 4B & 72.5 & 63.2 & 54.4 & 70.8 & 64.7 & 53.3 & 77.5 & 64.0 & 58.3 & 72.4 & 63.3 & 55.9 & 72.9 & 62.5 & 78.8 & 73.6 & 69.6 & 77.3 & 72.0 & 79.2 & 70.7 & \textcolor{GrapeDelta}{-10.2} \\
\GrapeModel{}{1}{Qwen3.5} & 2B & 70.8 & 59.6 & 49.7 & 67.3 & 55.7 & 48.0 & 76.0 & 60.9 & 50.0 & 73.9 & 58.1 & 52.1 & 65.5 & 53.4 & 72.0 & 66.3 & 60.1 & 68.9 & 62.9 & 72.3 & 63.8 & \textcolor{GrapeDelta}{-11.3} \\
\GrapeModel{}{1}{Qwen3.5} & 0.8B & 66.9 & 44.3 & 41.1 & 58.4 & 46.8 & 40.0 & 70.3 & 49.2 & 42.8 & 67.1 & 46.4 & 39.6 & 57.5 & 48.7 & 63.7 & 58.3 & 54.3 & 59.4 & 57.0 & 63.8 & 56.0 & \textcolor{GrapeDelta}{-10.4} \\
\GrapeModel{\GrapeBadge{qwen}}{0}{Qwen3-VL} & 8B & 73.1 & 62.4 & 56.7 & 69.5 & 62.3 & 57.5 & 77.3 & 60.9 & 56.8 & 75.5 & 63.3 & 56.9 & 72.1 & 61.6 & 77.5 & 72.5 & 68.9 & 76.1 & 71.8 & 78.8 & 70.2 & \textcolor{GrapeDelta}{-9.6} \\
\GrapeModel{}{1}{Qwen3-VL} & 4B & 71.1 & 56.1 & 50.5 & 68.9 & 62.1 & 55.5 & 73.7 & 57.7 & 54.7 & 71.3 & 59.7 & 52.7 & 69.8 & 60.5 & 76.5 & 70.3 & 67.1 & 73.8 & 69.7 & 76.1 & 67.9 & \textcolor{GrapeDelta}{-10.1} \\
\GrapeModel{\GrapeBadge{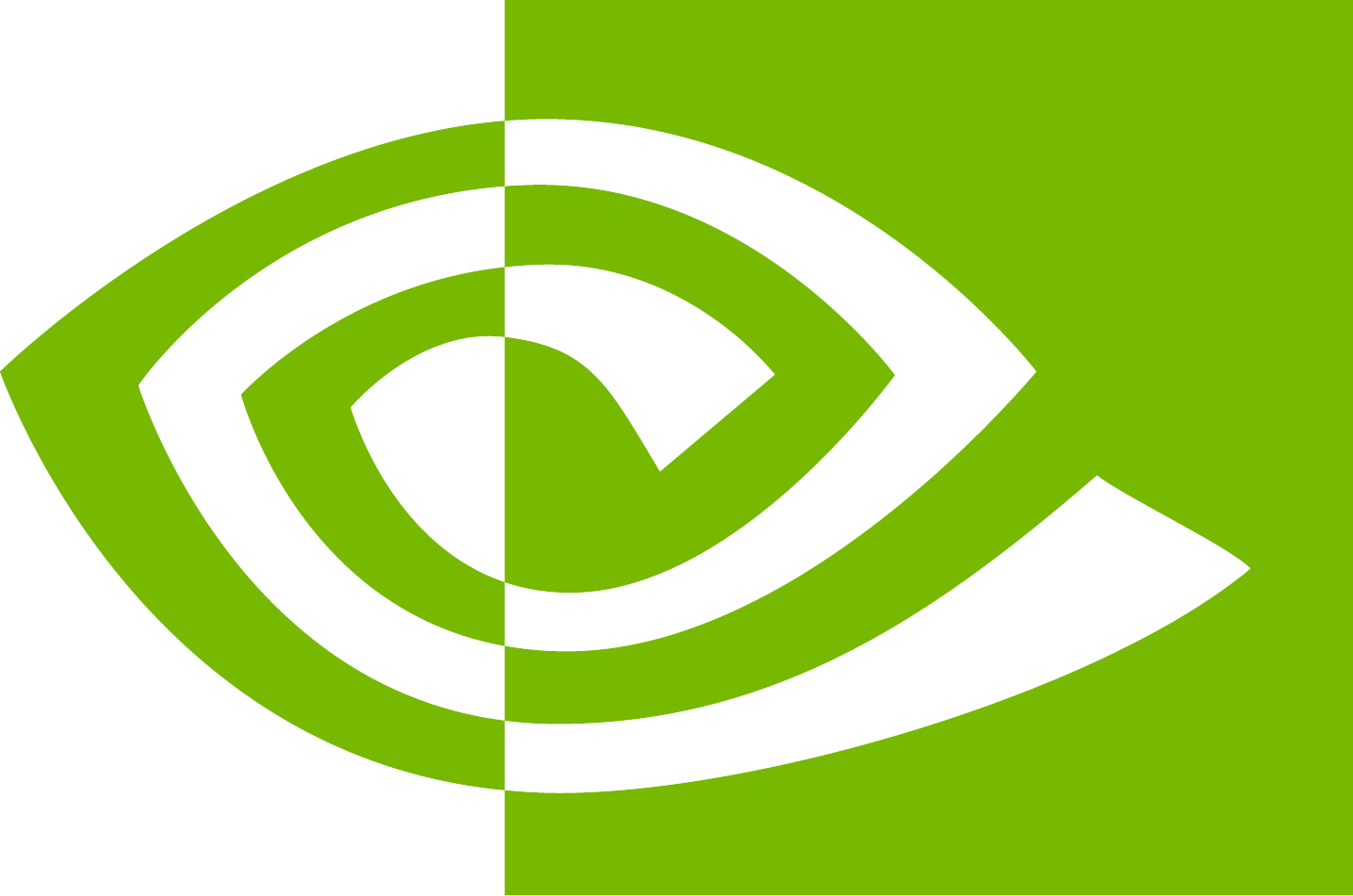}}{0}{Cosmos-Reason2} & 32B & 73.2 & 64.0 & 58.5 & 72.1 & 64.4 & 59.6 & 77.1 & 64.1 & 62.4 & 75.1 & 62.9 & 60.3 & 73.9 & 64.8 & 79.3 & 73.7 & 71.9 & 78.0 & 74.1 & 79.4 & 72.1 & \textcolor{GrapeDelta}{-10.0} \\
\GrapeModel{}{1}{Cosmos-Reason2} & 8B & 73.9 & 56.7 & 51.7 & 68.1 & 57.5 & 50.8 & 76.8 & 59.2 & 53.3 & 77.1 & 59.2 & 51.9 & 69.7 & 59.8 & 74.8 & 69.1 & 66.6 & 72.8 & 67.7 & 74.0 & 67.1 & \textcolor{GrapeDelta}{-9.8} \\
\GrapeModel{\GrapeBadge{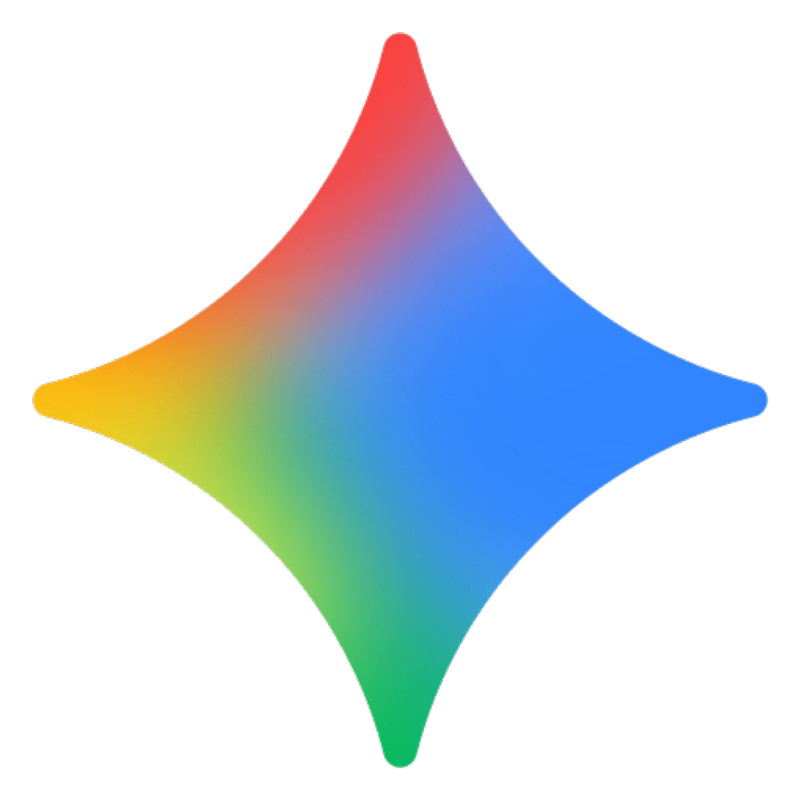}}{0}{Gemma 4} & 31B & 79.5 & 70.1 & 62.9 & 72.8 & \GrapeThird{GrapeDriving}{71.2} & 62.4 & 82.8 & 73.1 & 68.3 & 78.8 & \GrapeThird{GrapeDriving}{73.6} & 66.9 & 77.4 & 68.8 & 80.5 & 76.6 & 74.6 & 77.9 & 74.7 & 80.5 & 75.1 & \GrapeDeltaThird{-8.1} \\
\GrapeModel{\GrapeBadge{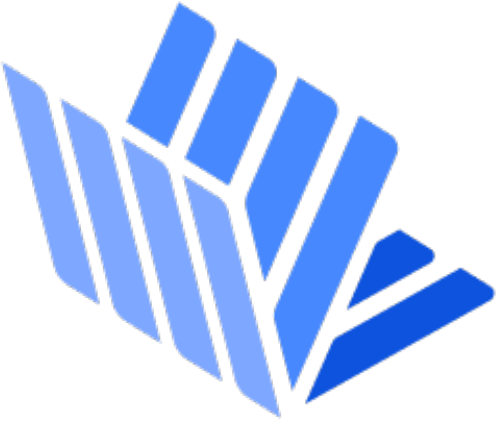}}{0}{InternVL3.5} & 38B & 80.3 & 66.7 & 62.9 & 76.0 & 68.4 & 61.2 & 82.9 & 67.2 & 63.9 & \GrapeSecond{GrapeDriving}{82.3} & 71.3 & 62.8 & 77.5 & 64.0 & 80.9 & 76.6 & 72.5 & 76.7 & 74.2 & 80.9 & 74.0 & \textcolor{GrapeDelta}{-8.5} \\
\GrapeModel{}{1}{InternVL3.5} & 14B & 73.1 & 61.2 & 54.5 & 66.4 & 60.1 & 56.0 & 74.9 & 62.8 & 58.8 & 71.7 & 63.2 & 57.9 & 71.9 & 58.6 & 75.4 & 72.5 & 67.6 & 73.2 & 69.0 & 78.3 & 68.8 & \textcolor{GrapeDelta}{-10.4} \\
\GrapeModel{}{1}{InternVL3.5} & 8B & 72.5 & 57.6 & 53.5 & 69.7 & 58.7 & 49.5 & 77.5 & 60.4 & 53.3 & 70.7 & 60.8 & 51.3 & 71.4 & 57.6 & 74.4 & 70.3 & 64.9 & 72.1 & 68.6 & 76.4 & 67.2 & \textcolor{GrapeDelta}{-10.7} \\
\GrapeModel{}{1}{InternVL3.5} & 4B & 69.3 & 52.8 & 49.2 & 64.5 & 51.3 & 44.8 & 75.5 & 53.9 & 53.3 & 71.2 & 56.1 & 49.3 & 68.2 & 53.8 & 72.0 & 67.8 & 62.2 & 69.5 & 65.3 & 73.3 & 64.0 & \textcolor{GrapeDelta}{-10.9} \\
\GrapeModel{\GrapeBadge{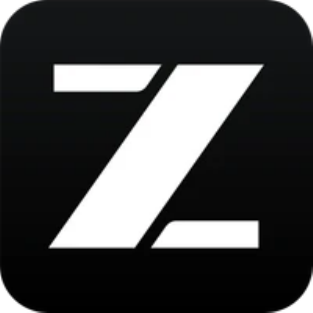}}{0}{GLM-4.1V-Thinking} & 9B & 68.0 & 58.0 & 53.7 & 58.3 & 57.3 & 53.6 & 71.3 & 58.7 & 55.6 & 64.1 & 55.7 & 52.0 & 68.1 & 56.1 & 71.5 & 62.8 & 65.5 & 69.8 & 66.8 & 72.6 & 64.5 & \textcolor{GrapeDelta}{-11.2} \\
\GrapeModel{\GrapeBadge{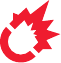}}{0}{MiniCPM-V 4.5} & 8B & 74.1 & 63.1 & 55.6 & 70.3 & 60.4 & 56.8 & 78.1 & 62.8 & 55.1 & 74.1 & 61.2 & 54.5 & 74.8 & 62.4 & 76.1 & 71.4 & 67.9 & 73.5 & 69.2 & 77.3 & 69.4 & \textcolor{GrapeDelta}{-9.7} \\
\GrapeModel{\GrapeBadge{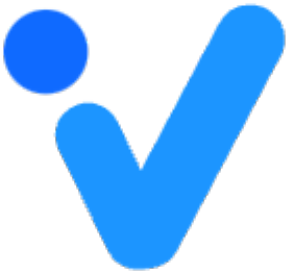}}{0}{Ovis2.5} & 9B & 73.6 & 54.7 & 51.5 & 68.1 & 52.7 & 50.9 & 76.3 & 56.0 & 55.6 & 73.5 & 53.2 & 47.7 & 68.7 & 56.0 & 73.6 & 70.0 & 63.7 & 69.3 & 66.5 & 73.1 & 65.4 & \textcolor{GrapeDelta}{-9.7} \\
\GrapeModel{}{1}{Ovis2.5} & 2B & 67.6 & 44.5 & 38.0 & 63.2 & 41.7 & 35.9 & 70.5 & 43.2 & 42.1 & 67.1 & 42.1 & 35.5 & 58.6 & 47.6 & 61.9 & 59.5 & 50.5 & 58.5 & 56.5 & 63.7 & 54.9 & \textcolor{GrapeDelta}{-10.4} \\
\GrapeModel{\GrapeBadge{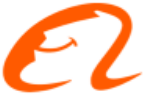}}{0}{VideoLLaMA3} & 7B & 63.2 & 46.9 & 52.7 & 62.0 & 42.8 & 51.5 & 67.9 & 48.4 & 54.7 & 65.9 & 47.7 & 48.0 & 64.6 & 52.1 & 70.2 & 62.2 & 60.7 & 66.9 & 60.9 & 69.9 & 60.9 & \textcolor{GrapeDelta}{-11.4} \\
\GrapeModel{\GrapeBadge{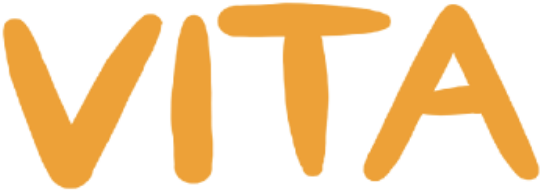}}{0}{VITA-1.5} & 7B & 60.4 & 50.0 & 47.3 & 58.0 & 51.1 & 47.9 & 68.0 & 52.7 & 50.5 & 59.7 & 47.9 & 46.0 & 61.7 & 53.8 & 66.7 & 60.5 & 58.3 & 63.3 & 59.1 & 66.8 & 59.1 & \textcolor{GrapeDelta}{-9.4} \\
\arrayrulecolor{GrapeText}\bottomrule[.8pt]
\end{tabular}}
\typeout{GRAPE-main-WIDTH=\the\wd\GrapeMeasureBox}
\typeout{GRAPE-main-HEIGHT=\the\dimexpr\ht\GrapeMeasureBox+\dp\GrapeMeasureBox\relax}
\noindent\resizebox{\linewidth}{!}{\usebox{\GrapeMeasureBox}}
\endgroup
\endgroup

\end{minipage}
\vspace{-10pt}
\end{table*}
\begin{table*}[t]
\centering
\begin{minipage}{\textwidth}
\caption{\textbf{Results on the \grapebench{} real-world set.} Accuracy (\%; higher is better) on driving videos captured under naturally occurring rain, snow, nighttime, and fog conditions. L1--L3 are reported separately for each condition, and each condition's Avg. pools its questions across all levels. Rank orders models by decreasing All Avg. (1 is best). Rank and top-3 shading are computed separately for proprietary and open-source VLMs. Fine-tuned variants use Qwen3.5-9B and the synthetic training split only (Section~\ref{sec:exp}). }
\vspace{-5pt}
\label{tab:grape_realworld_levels}
\begingroup
\setlength{\tabcolsep}{5.5pt}
\renewcommand{\arraystretch}{1.05}
\fontsize{8}{9.5}\selectfont
\color{GrapeText}
\newcommand{\GrapeRealHead}[2]{\raisebox{\dimexpr 2.7pt-\height/2\relax}[0pt][0pt]{\GrapeIcon{GrapeDriving}{#1}}\hspace{3pt}\textcolor{GrapeDriving}{\textbf{#2}}}
\newcommand{\GrapeTuneTag}[1]{\hspace{.6pt}\raisebox{-1.1pt}[0pt][0pt]{\fontsize{6}{6.5}\selectfont\itshape #1}}
\sbox{\GrapeMeasureBox}{%
\begin{tabular}{@{}l c c *{20}{>{\centering\arraybackslash}p{15pt}}}
\arrayrulecolor{GrapeText}\toprule[.8pt]
\multicolumn{1}{l}{\multirow{2}{*}{\textbf{Model}}} & \multirow{2}{*}{\textbf{Size}} & \multirow{2}{*}{\textbf{Rank}} & \multicolumn{4}{c}{\GrapeRealHead{Rain}{Rain}} & \multicolumn{4}{c}{\GrapeRealHead{Snow}{Snow}} & \multicolumn{4}{c}{\GrapeRealHead{Night}{Night}} & \multicolumn{4}{c}{\GrapeRealHead{Fog}{Fog}} & \multicolumn{4}{c}{\textcolor{GrapeAvg!75!black}{\textbf{All}}} \\
\arrayrulecolor{GrapeRule}\cmidrule(lr){4-7}\cmidrule(lr){8-11}\cmidrule(lr){12-15}\cmidrule(lr){16-19}\cmidrule(lr){20-23}
 & & & L1 & L2 & L3 & \textcolor{GrapeDriving}{\textbf{Avg.}} & L1 & L2 & L3 & \textcolor{GrapeDriving}{\textbf{Avg.}} & L1 & L2 & L3 & \textcolor{GrapeDriving}{\textbf{Avg.}} & L1 & L2 & L3 & \textcolor{GrapeDriving}{\textbf{Avg.}} & \textcolor{GrapeAvg!75!black}{L1} & \textcolor{GrapeAvg!75!black}{L2} & \textcolor{GrapeAvg!75!black}{L3} & \textcolor{GrapeAvg!75!black}{\textbf{Avg.}} \\
\arrayrulecolor{GrapeText}\midrule[.5pt]
\GrapeSection{23}{Proprietary VLMs}
\GrapeModel{\GrapeBadge{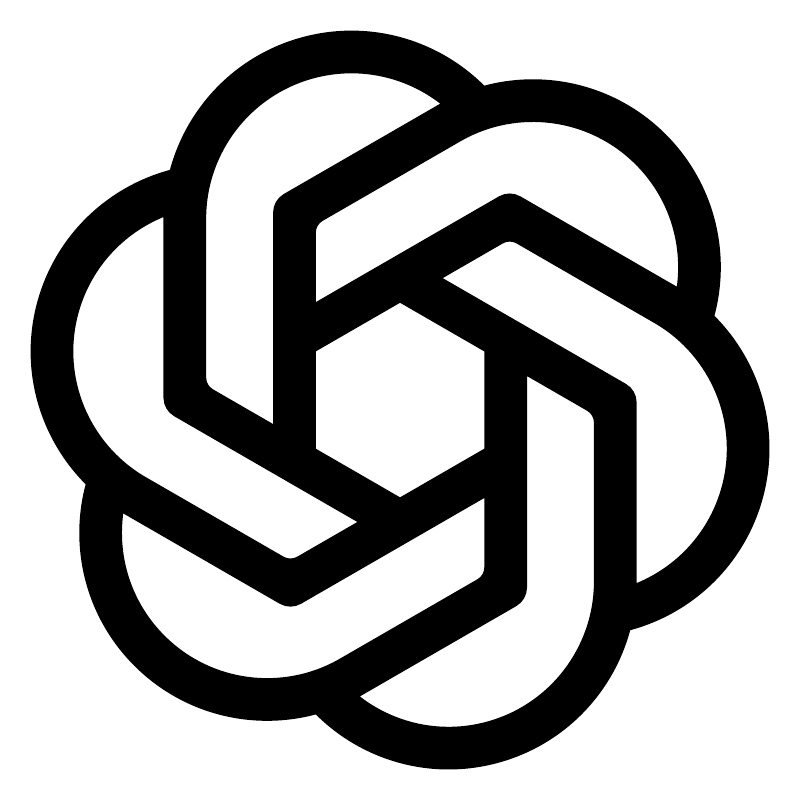}}{0}{GPT-6 Luna} & -- & 3 & \GrapeThird{GrapeDriving}{84.7} & \GrapeSecond{GrapeDriving}{87.4} & \GrapeSecond{GrapeDriving}{82.1} & \GrapeSecond{GrapeDriving}{84.7} & 83.5 & \GrapeThird{GrapeDriving}{86.2} & 69.7 & 79.8 & \GrapeThird{GrapeDriving}{88.9} & \GrapeThird{GrapeDriving}{85.0} & \GrapeThird{GrapeDriving}{75.8} & \GrapeThird{GrapeDriving}{83.2} & \GrapeFirst{GrapeDriving}{96.0} & \GrapeSecond{GrapeDriving}{93.0} & 91.0 & \GrapeThird{GrapeDriving}{93.3} & \GrapeThird{GrapeAvg}{87.9} & \GrapeThird{GrapeAvg}{87.8} & \GrapeThird{GrapeAvg}{79.5} & \GrapeThird{GrapeAvg}{85.1} \\
\GrapeModel{\GrapeBadge{openai}}{0}{GPT-5.6 Sol} & -- & 1 & \GrapeFirst{GrapeDriving}{85.3} & \GrapeFirst{GrapeDriving}{90.5} & \GrapeFirst{GrapeDriving}{83.2} & \GrapeFirst{GrapeDriving}{86.3} & \GrapeFirst{GrapeDriving}{91.7} & \GrapeSecond{GrapeDriving}{89.0} & \GrapeFirst{GrapeDriving}{79.8} & \GrapeFirst{GrapeDriving}{86.9} & \GrapeFirst{GrapeDriving}{91.5} & \GrapeSecond{GrapeDriving}{85.6} & \GrapeFirst{GrapeDriving}{82.4} & \GrapeFirst{GrapeDriving}{86.5} & \GrapeThird{GrapeDriving}{94.0} & \GrapeFirst{GrapeDriving}{96.0} & \GrapeSecond{GrapeDriving}{92.0} & \GrapeFirst{GrapeDriving}{94.0} & \GrapeFirst{GrapeAvg}{90.2} & \GrapeFirst{GrapeAvg}{89.9} & \GrapeFirst{GrapeAvg}{83.4} & \GrapeFirst{GrapeAvg}{87.8} \\
\GrapeModel{\GrapeBadge{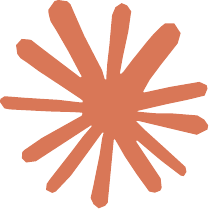}}{0}{Claude Opus 5.5} & -- & 2 & \GrapeFirst{GrapeDriving}{85.3} & \GrapeThird{GrapeDriving}{86.8} & \GrapeThird{GrapeDriving}{80.5} & \GrapeThird{GrapeDriving}{84.2} & \GrapeSecond{GrapeDriving}{90.8} & \GrapeFirst{GrapeDriving}{89.9} & \GrapeThird{GrapeDriving}{76.1} & \GrapeSecond{GrapeDriving}{85.6} & \GrapeSecond{GrapeDriving}{89.5} & \GrapeFirst{GrapeDriving}{88.2} & \GrapeSecond{GrapeDriving}{76.5} & \GrapeSecond{GrapeDriving}{84.7} & 93.0 & \GrapeSecond{GrapeDriving}{93.0} & \GrapeFirst{GrapeDriving}{95.0} & \GrapeSecond{GrapeDriving}{93.7} & \GrapeSecond{GrapeAvg}{89.2} & \GrapeSecond{GrapeAvg}{89.0} & \GrapeSecond{GrapeAvg}{80.9} & \GrapeSecond{GrapeAvg}{86.4} \\
\GrapeModel{\GrapeBadge{claude}}{0}{Claude Sonnet 5.5} & -- & 4 & 83.2 & 80.5 & 78.9 & 80.9 & \GrapeThird{GrapeDriving}{88.1} & 85.3 & \GrapeSecond{GrapeDriving}{77.1} & \GrapeThird{GrapeDriving}{83.5} & 85.0 & 83.7 & 71.9 & 80.2 & \GrapeSecond{GrapeDriving}{95.0} & 91.0 & \GrapeSecond{GrapeDriving}{92.0} & 92.7 & 87.1 & 84.4 & 78.3 & 83.3 \\
\GrapeModel{\GrapeBadge{claude}}{0}{Claude Haiku 4.5} & -- & 6 & 61.6 & 61.1 & 54.7 & 59.1 & 70.6 & 63.3 & 51.4 & 61.8 & 63.4 & 60.8 & 60.1 & 61.4 & 75.0 & 72.0 & 70.0 & 72.3 & 66.1 & 63.5 & 58.7 & 62.8 \\
\GrapeModel{\GrapeBadge{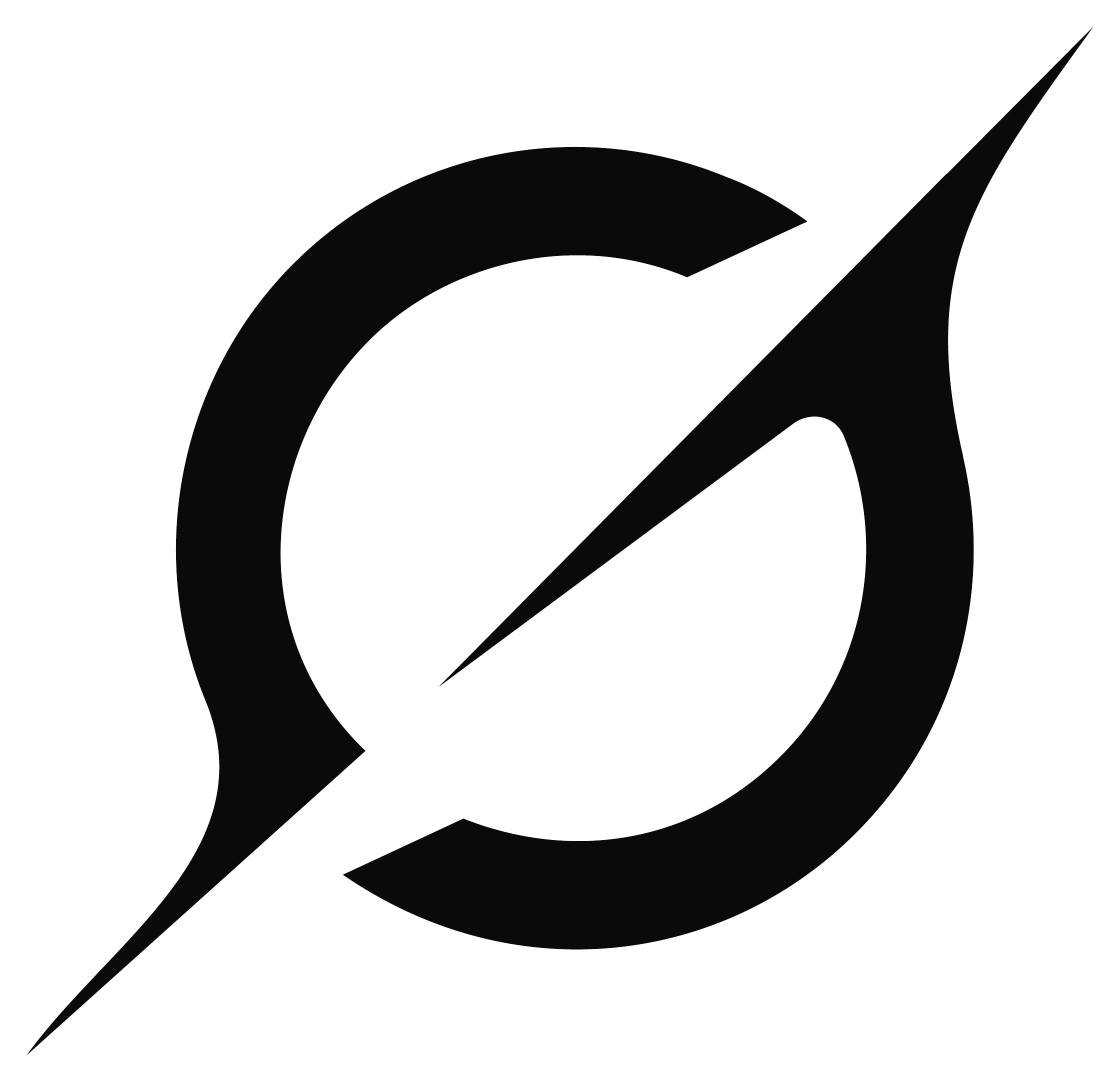}}{0}{Grok 4.7} & -- & 5 & 73.7 & 77.9 & 74.2 & 75.3 & 79.8 & 81.7 & 62.4 & 74.6 & 83.0 & 81.0 & 68.0 & 77.3 & 93.0 & 91.0 & 89.0 & 91.0 & 81.5 & 81.8 & 72.6 & 78.6 \\
\midrule
\GrapeSection{23}{Open-Source VLMs}
\GrapeModel{\GrapeBadge{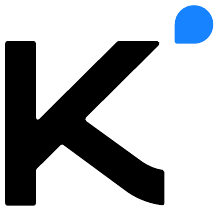}}{0}{Kimi K3} & 2.8T & 1 & \GrapeSecond{GrapeDriving}{83.7} & \GrapeFirst{GrapeDriving}{87.9} & \GrapeFirst{GrapeDriving}{79.5} & \GrapeFirst{GrapeDriving}{83.7} & \GrapeSecond{GrapeDriving}{88.1} & \GrapeFirst{GrapeDriving}{89.9} & \GrapeSecond{GrapeDriving}{71.6} & \GrapeFirst{GrapeDriving}{83.2} & 84.3 & \GrapeFirst{GrapeDriving}{87.6} & \GrapeFirst{GrapeDriving}{79.7} & \GrapeFirst{GrapeDriving}{83.9} & \GrapeSecond{GrapeDriving}{92.0} & \GrapeFirst{GrapeDriving}{92.0} & \GrapeFirst{GrapeDriving}{95.0} & \GrapeFirst{GrapeDriving}{93.0} & \GrapeSecond{GrapeAvg}{86.7} & \GrapeFirst{GrapeAvg}{88.8} & \GrapeFirst{GrapeAvg}{80.8} & \GrapeFirst{GrapeAvg}{85.4} \\
\GrapeModel{\GrapeBadge{qwen}}{0}{Qwen3.8-Flash-Next} & 125B & 2 & \GrapeFirst{GrapeDriving}{84.7} & \GrapeThird{GrapeDriving}{82.6} & \GrapeSecond{GrapeDriving}{78.4} & \GrapeSecond{GrapeDriving}{81.9} & \GrapeFirst{GrapeDriving}{90.8} & 83.5 & \GrapeThird{GrapeDriving}{70.6} & \GrapeSecond{GrapeDriving}{81.7} & \GrapeFirst{GrapeDriving}{90.8} & 78.4 & \GrapeSecond{GrapeDriving}{77.8} & \GrapeSecond{GrapeDriving}{82.4} & \GrapeFirst{GrapeDriving}{94.0} & \GrapeSecond{GrapeDriving}{88.0} & 86.0 & \GrapeThird{GrapeDriving}{89.3} & \GrapeFirst{GrapeAvg}{89.5} & 82.5 & \GrapeSecond{GrapeAvg}{77.6} & \GrapeSecond{GrapeAvg}{83.2} \\
\GrapeModel{\GrapeBadge{qwen}}{0}{Qwen3.8} & 27B & 4 & \GrapeThird{GrapeDriving}{82.6} & \GrapeSecond{GrapeDriving}{83.7} & \GrapeThird{GrapeDriving}{74.2} & \GrapeThird{GrapeDriving}{80.2} & \GrapeThird{GrapeDriving}{86.2} & 79.8 & 67.0 & 77.7 & \GrapeThird{GrapeDriving}{87.6} & 79.1 & 73.9 & 80.2 & \GrapeSecond{GrapeDriving}{92.0} & 84.0 & 89.0 & 88.3 & \GrapeSecond{GrapeAvg}{86.7} & 82.0 & 75.0 & 81.2 \\
\GrapeModel{\GrapeBadge{qwen}}{0}{Qwen3.6} & 27B & 5 & 82.1 & 79.5 & 70.5 & 77.4 & \GrapeThird{GrapeDriving}{86.2} & \GrapeSecond{GrapeDriving}{84.4} & 67.9 & 79.5 & \GrapeThird{GrapeDriving}{87.6} & \GrapeThird{GrapeDriving}{81.7} & 74.5 & 81.3 & 86.0 & \GrapeSecond{GrapeDriving}{88.0} & \GrapeThird{GrapeDriving}{90.0} & 88.0 & 85.5 & \GrapeThird{GrapeAvg}{82.9} & 74.1 & 80.8 \\
\GrapeModel{\GrapeBadge{qwen}}{0}{Qwen3.5} & 27B & 3 & 80.0 & 82.1 & 73.7 & 78.6 & \GrapeThird{GrapeDriving}{86.2} & \GrapeSecond{GrapeDriving}{84.4} & \GrapeFirst{GrapeDriving}{72.5} & \GrapeThird{GrapeDriving}{81.0} & \GrapeThird{GrapeDriving}{87.6} & \GrapeSecond{GrapeDriving}{82.4} & \GrapeThird{GrapeDriving}{76.5} & \GrapeThird{GrapeDriving}{82.1} & 91.0 & \GrapeSecond{GrapeDriving}{88.0} & \GrapeSecond{GrapeDriving}{92.0} & \GrapeSecond{GrapeDriving}{90.3} & 85.7 & \GrapeSecond{GrapeAvg}{83.7} & \GrapeThird{GrapeAvg}{77.1} & \GrapeThird{GrapeAvg}{82.2} \\
\GrapeModel{}{1}{Qwen3.5} & 9B & 7 & 76.3 & 74.7 & 70.0 & 73.7 & 85.3 & 79.8 & 59.6 & 74.9 & 85.0 & 75.2 & 71.9 & 77.3 & 84.0 & 86.0 & 82.0 & 84.0 & 82.0 & 78.3 & 69.9 & 76.7 \\
\GrapeModel{}{1}{Qwen3.5} & 4B & 9 & 75.3 & 75.3 & 68.9 & 73.2 & 82.6 & 81.7 & 56.9 & 73.7 & 81.7 & 75.8 & 74.5 & 77.3 & 80.0 & 83.0 & 77.0 & 80.0 & 79.5 & 78.1 & 69.4 & 75.7 \\
\GrapeModel{}{1}{Qwen3.5} & 2B & 12 & 74.7 & 72.1 & 65.3 & 70.7 & 79.8 & 66.1 & 47.7 & 64.5 & 85.0 & 70.6 & 69.9 & 75.2 & 82.0 & 78.0 & 81.0 & 80.3 & 80.1 & 71.7 & 65.4 & 72.4 \\
\GrapeModel{}{1}{Qwen3.5} & 0.8B & 20 & 66.3 & 64.7 & 60.0 & 63.7 & 78.0 & 63.3 & 41.3 & 60.9 & 73.2 & 68.0 & 68.0 & 69.7 & 76.0 & 65.0 & 75.0 & 72.0 & 71.9 & 65.0 & 60.3 & 65.7 \\
\GrapeModel{\GrapeBadge{qwen}}{0}{Qwen3-VL} & 8B & 13 & 70.5 & 66.8 & 64.7 & 67.4 & 79.8 & 71.6 & 57.8 & 69.7 & 81.7 & 67.3 & 66.7 & 71.9 & 84.0 & 82.0 & 83.0 & 83.0 & 78.0 & 70.1 & 67.0 & 71.7 \\
\GrapeModel{}{1}{Qwen3-VL} & 4B & 15 & 68.4 & 65.3 & 64.2 & 66.0 & 75.2 & 67.0 & 48.6 & 63.6 & 75.8 & 73.2 & 66.7 & 71.9 & 82.0 & 81.0 & 80.0 & 81.0 & 74.5 & 70.6 & 64.3 & 69.8 \\
\GrapeModel{\GrapeBadge{cosmos}}{0}{Cosmos-Reason2} & 32B & 6 & 78.4 & 78.4 & \GrapeThird{GrapeDriving}{74.2} & 77.0 & \GrapeThird{GrapeDriving}{86.2} & 82.6 & 64.2 & 77.7 & \GrapeSecond{GrapeDriving}{88.2} & 75.8 & \GrapeThird{GrapeDriving}{76.5} & 80.2 & 89.0 & 86.0 & 87.0 & 87.3 & 85.0 & 79.9 & 74.8 & 79.9 \\
\GrapeModel{}{1}{Cosmos-Reason2} & 8B & 8 & 74.7 & 77.4 & 70.5 & 74.2 & 83.5 & 75.2 & 56.9 & 71.9 & 85.6 & 71.2 & 66.0 & 74.3 & 89.0 & 87.0 & 83.0 & 86.3 & 82.5 & 77.1 & 68.4 & 76.0 \\
\GrapeModel{\GrapeBadge{google-sparkle}}{0}{Gemma 4} & 31B & 10 & 68.9 & 72.1 & 67.9 & 69.6 & 79.8 & 76.1 & 61.5 & 72.5 & 79.7 & 70.6 & 69.3 & 73.2 & 88.0 & 80.0 & 78.0 & 82.0 & 77.6 & 74.1 & 69.1 & 73.6 \\
\GrapeModel{\GrapeBadge{internvl}}{0}{InternVL3.5} & 38B & 11 & 70.0 & 70.5 & 67.9 & 69.5 & 82.6 & 69.7 & 61.5 & 71.3 & 76.5 & 73.2 & 70.6 & 73.4 & 86.0 & 80.0 & 81.0 & 82.3 & 77.3 & 72.6 & 68.9 & 72.9 \\
\GrapeModel{}{1}{InternVL3.5} & 14B & 19 & 67.9 & 69.5 & 59.5 & 65.6 & 77.1 & 64.2 & 48.6 & 63.3 & 71.2 & 69.3 & 62.7 & 67.8 & 82.0 & 78.0 & 72.0 & 77.3 & 73.3 & 69.4 & 60.1 & 67.6 \\
\GrapeModel{}{1}{InternVL3.5} & 8B & 17 & 67.4 & 62.6 & 65.3 & 65.1 & 76.1 & 67.0 & 55.0 & 66.1 & 74.5 & 67.3 & 60.8 & 67.5 & 82.0 & 77.0 & 77.0 & 78.7 & 73.6 & 67.5 & 64.2 & 68.4 \\
\GrapeModel{}{1}{InternVL3.5} & 4B & 23 & 64.7 & 59.5 & 54.7 & 59.6 & 76.1 & 62.4 & 56.0 & 64.8 & 69.9 & 60.8 & 62.1 & 64.3 & 77.0 & 77.0 & 76.0 & 76.7 & 70.3 & 63.5 & 60.7 & 64.8 \\
\GrapeModel{\GrapeBadge{zai}}{0}{GLM-4.1V-Thinking} & 9B & 14 & 66.8 & 70.0 & 68.9 & 68.6 & 80.7 & 69.7 & 56.0 & 68.8 & 75.2 & 72.5 & 65.4 & 71.0 & 82.0 & 82.0 & 73.0 & 79.0 & 74.8 & 72.6 & 65.6 & 71.0 \\
\GrapeModel{\GrapeBadge{minicpm}}{0}{MiniCPM-V 4.5} & 8B & 16 & 65.3 & 68.4 & 70.0 & 67.9 & 77.1 & 67.0 & 58.7 & 67.6 & 75.2 & 64.7 & 68.0 & 69.3 & 79.0 & 76.0 & 68.0 & 74.3 & 72.9 & 68.4 & 66.4 & 69.2 \\
\GrapeModel{\GrapeBadge{ovis}}{0}{Ovis2.5} & 9B & 21 & 63.2 & 61.1 & 58.9 & 61.1 & 73.4 & 61.5 & 53.2 & 62.7 & 73.9 & 62.1 & 61.4 & 65.8 & 84.0 & 69.0 & 76.0 & 76.3 & 71.9 & 62.8 & 61.0 & 65.2 \\
\GrapeModel{}{1}{Ovis2.5} & 2B & 24 & 57.4 & 51.6 & 53.2 & 54.0 & 58.7 & 47.7 & 41.3 & 49.2 & 61.4 & 55.6 & 51.6 & 56.2 & 74.0 & 67.0 & 61.0 & 67.3 & 62.1 & 54.5 & 52.1 & 56.2 \\
\GrapeModel{\GrapeBadge{alibaba}}{0}{VideoLLaMA3} & 7B & 18 & 68.4 & 67.9 & 65.8 & 67.4 & 74.3 & 72.5 & 50.5 & 65.7 & 73.2 & 62.7 & 64.1 & 66.7 & 83.0 & 72.0 & 73.0 & 76.0 & 73.8 & 67.7 & 63.1 & 68.2 \\
\GrapeModel{\GrapeBadge{vita}}{0}{VITA-1.5} & 7B & 22 & 62.1 & 64.2 & 60.0 & 62.1 & 77.1 & 70.6 & 51.4 & 66.4 & 67.3 & 60.1 & 64.1 & 63.8 & 80.0 & 70.0 & 63.0 & 71.0 & 69.9 & 65.0 & 59.6 & 64.9 \\
\midrule
\GrapeSection{23}{Fine-Tuned VLMs}
\GrapeModel{\GrapeBadge{qwen}}{1}{Qwen3.5} & \textcolor{GrapeMuted}{9B} & \textcolor{GrapeMuted}{--} & \textcolor{GrapeMuted}{76.3} & \textcolor{GrapeMuted}{74.7} & \textcolor{GrapeMuted}{70.0} & \textcolor{GrapeMuted}{73.7} & \textcolor{GrapeMuted}{85.3} & \textcolor{GrapeMuted}{79.8} & \textcolor{GrapeMuted}{59.6} & \textcolor{GrapeMuted}{74.9} & \textcolor{GrapeMuted}{85.0} & \textcolor{GrapeMuted}{75.2} & \textcolor{GrapeMuted}{71.9} & \textcolor{GrapeMuted}{77.3} & \textcolor{GrapeMuted}{84.0} & \textcolor{GrapeMuted}{86.0} & \textcolor{GrapeMuted}{82.0} & \textcolor{GrapeMuted}{84.0} & \textcolor{GrapeMuted}{82.0} & \textcolor{GrapeMuted}{78.3} & \textcolor{GrapeMuted}{69.9} & \textcolor{GrapeMuted}{76.7} \\
\GrapeModel{}{0}{+ SFT\GrapeTuneTag{w/o reasoning}} & 9B & -- & 82.1 & 80.4 & 75.1 & 79.2 & 88.4 & 81.7 & 67.0 & 79.0 & 86.9 & 80.0 & 78.6 & 81.8 & 91.0 & 86.3 & 83.0 & 86.8 & 86.4 & 81.6 & 75.4 & 81.1 \\
\GrapeModel{}{0}{+ SFT\GrapeTuneTag{w/ reasoning}} & 9B & -- & 75.8 & 77.9 & 70.0 & 74.6 & 86.2 & 78.9 & 58.7 & 74.6 & 85.6 & 77.1 & 77.1 & 80.0 & 84.0 & 86.0 & 88.0 & 86.0 & 82.0 & 79.5 & 72.9 & 78.1 \\
\GrapeModel{}{0}{+ OPSD\GrapeTuneTag{w/ reasoning}} & 9B & -- & 81.6 & 77.9 & 72.1 & 77.2 & 89.9 & 79.8 & 67.0 & 78.9 & 89.5 & 79.1 & 76.5 & 81.7 & 89.0 & 84.0 & 86.0 & 86.3 & 86.9 & 80.1 & 74.1 & 80.4 \\
\arrayrulecolor{GrapeText}\bottomrule[.8pt]
\end{tabular}}
\typeout{GRAPE-real-WIDTH=\the\wd\GrapeMeasureBox}
\noindent\resizebox{\linewidth}{!}{\usebox{\GrapeMeasureBox}}
\endgroup
\end{minipage}
\vspace{-10pt}
\end{table*}

\endgroup

\grapebench{} evaluates video understanding across visual degradation and reasoning difficulty. It contains a synthetic set with twelve corruption types applied to daily-life and driving videos, and a real-world driving set with naturally occurring rain, snow, nighttime, and fog conditions. Both partitions organize questions into three levels (L1--L3), designed to require progressively more demanding reasoning.

\subsection{Task Formulation}
\label{sec:task-formulation}
We formulate the task as multiple-choice question answering on visually degraded videos. Given a video $v$, question $q$, and candidate answer set $\mathcal{A}$, a model selects an option and is evaluated against the reference answer $y$. Each video context is paired with three questions (L1--L3), answered independently without answers or rationales from other levels. L1 targets a coarse entity or event, L2 a focused property, state, action, or relation, and L3 the integration of relevant facts with additional visual evidence. These levels specify intended reasoning demands; their empirical difficulty is assessed through level-wise accuracy.
Synthetic samples pair a clean video $v^0$ with a corrupted counterpart $\tilde{v}$, keeping questions, options, and reference answers identical. Real-world samples use naturally degraded driving videos. Appendix~\ref{app:evaluation-protocol} details the video input protocol.

\subsection{Benchmark Construction}
\label{sec:data-construction}

\noindent\textbf{Data source}.
The synthetic set covers driving scenarios from
VRU-Accident~\cite{kim2025vruaccident}, alongside general video
understanding and daily-life scenarios from
Video-MME~\cite{fu2025videomme}, STAR~\cite{wu2021star},
NExT-QA~\cite{xiao2021nextqa}, and
EgoSchema~\cite{mangalam2023egoschema}.
The source QA annotations used for synthetic construction follow
a multiple-choice format.
The real-world set focuses on driving scenarios, drawing videos
captured under naturally adverse conditions from
DrivingDojo~\cite{wang2024drivingdojo},
Boreas~\cite{burnett2023boreas},
Boreas-RT~\cite{lisus2026boreasrt},
BDD100K~\cite{yu2020bdd100k}, and the DrivingHazy and
GoProHazy collections~\cite{fan2024dvd}, for which annotators
construct and finalize multiple-choice questions.

\begin{figure}[t]
  \centering
  \includegraphics[width=\linewidth]{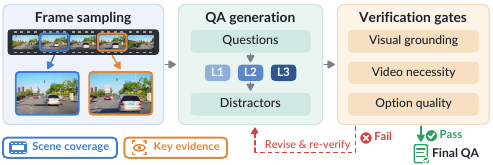}
  \vspace{-15pt}
  \caption{\textbf{Synthetic question construction in \grapebench{}.} Sampled frames provide scene context and question-relevant evidence for generating L1--L3 questions, reference answers, and distractors. Candidates undergo six verification gates.}
  \label{fig:data-construction}
  \vspace{-10pt}
\end{figure}

\vspace{3pt} \noindent
\textbf{Synthetic set construction.}
Figure~\ref{fig:data-construction} outlines three modules. \emph{(i) Frame sampling:} we sample source-video frames for scene context and question-relevant evidence. \emph{(ii) QA generation:} we generate L1--L3 questions and reference answers, then add distractors. \emph{(iii) Verification:} we apply six shared verification gates, including checks for visual grounding, video necessity, and distractor plausibility. Failed candidates are revised and re-verified, with additional evidence sampled when needed. See Appendix~\ref{app:synthetic-construction} for details. 
Each question group is paired with one corrupted version of its source video, preserving its questions, answer options, and reference answers. \looseness=-1

\vspace{3pt} \noindent
\textbf{Real-world set construction.}
We select driving clips captured under naturally adverse weather or nighttime conditions.
Question construction combines direct human authoring with AI-generated drafts that annotators review and revise before finalization.
Each retained clip contains one question at each of L1--L3, with four answer options and one reference answer per question, following the same high-level question hierarchy as the synthetic set.
Annotators inspect the video evidence when preparing the questions and exclude clips judged unsuitable for annotation.
Appendix~\ref{app:realworld-construction} details clip preparation and the annotation procedure.

\begin{figure}[tp]
  \centering
  \includegraphics[width=\columnwidth]{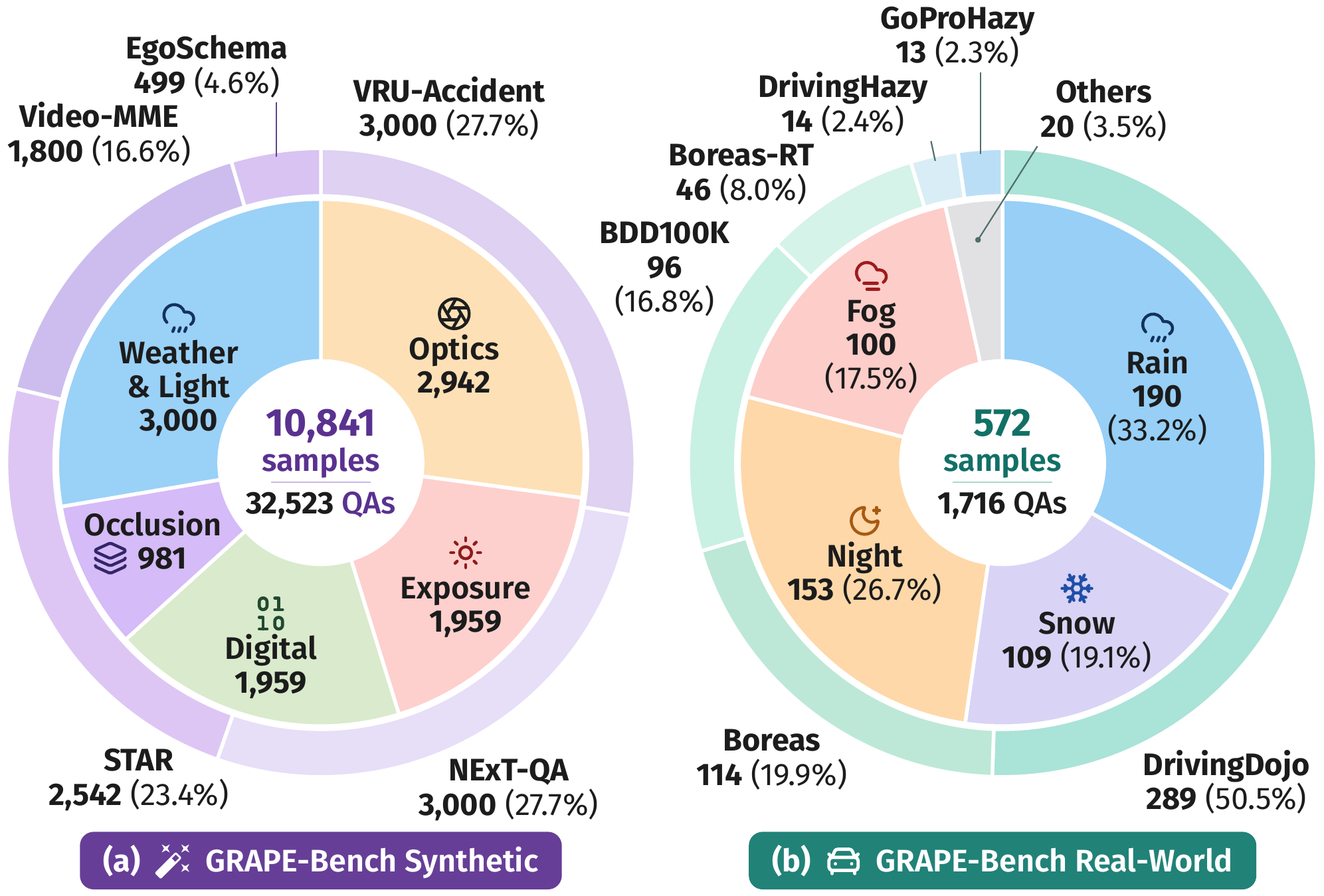}
  \vspace{-10pt}
  \caption{\textbf{Dataset statistics of \grapebench{}}. \textbf{(a)} Synthetic set with 10,841 corrupted-video samples and 32,523 questions. \textbf{(b)} Real-world set with 572 videos and 1,716 questions. The outer rings show source datasets, while the inner rings summarize corruption families. Each sample is associated with three questions of progressively increasing difficulty (L1–L3).}
  \label{fig:dataset-statistics}
  \vspace{-5pt}
\end{figure}

\subsection{Data Statistics}
\label{sec:data-statistics}

\noindent
\textbf{Synthetic set.}
The released synthetic inventory contains 10,841
video--question-group samples and 32,523 questions.
Of these, 10,818 groups (32,454 questions) passed automatic
verification. We manually reviewed, revised, and approved the remaining 23.
Figure~\ref{fig:dataset-statistics}(a) summarizes the released
synthetic inventory, also reported in Table~\ref{tab:benchmark-comparison}.
Samples are counted by question group: several groups can
originate from the same source video.
The four driving corruptions (rain, snow, nighttime, and fog)
each account for about 750 samples drawn from VRU-Accident,
while each of the eight daily-life corruptions, including
over- and underexposure, blur, and bitstream errors, accounts
for roughly 980 samples drawn from the other source datasets.
The corruption distribution is approximately balanced within
each scenario pool. Video-MME, VRU-Accident, STAR, and NExT-QA
together account for 95.4\% of all QA samples.

\vspace{3pt} \noindent
\textbf{Real-world set.}
The real-world set contains 572 videos and 1,716
questions. It includes 190 videos captured in rain, 109 in
snow, 153 at night, 100 in fog, and 20 labeled as \emph{others}.
Each video carries one question at each of L1--L3,
yielding 572 questions per level.
Figure~\ref{fig:dataset-statistics}(b) summarizes the source
and condition distributions.
The real-world overall score includes all 572 videos, including
\emph{others}; the four named conditions account for 552 videos
and 1,656 questions.

\subsection{Evaluation Results}
\label{sec:evaluation-protocol}

\noindent\textbf{Evaluation protocol.}
Tables~\ref{tab:grape_main_levels} and~\ref{tab:grape_realworld_levels}
report multiple-choice accuracy on the synthetic and real-world
sets, respectively.
Each question receives equal weight when computing accuracy.
For the synthetic set, the \emph{All} column reports accuracy across
all twelve corruptions and three difficulty levels.
The $\Delta$ column reports the change from clean to corrupted videos
on the same questions, in percentage points; negative values indicate
a loss in accuracy.
For the real-world set, the \emph{All} group covers all 572 videos,
including those labeled \emph{others}.
Its L1--L3 columns report accuracy at each difficulty level, while
\emph{Avg.} reports accuracy across all 1,716 questions.

\vspace{3pt} \noindent
\textbf{Evaluated models.} We evaluate 23 state-of-the-art open-weight models,
covering
Qwen~\cite{bai2025qwen3vl,qwen2026qwen35,qwen2026qwen36,
qwen2026qwen38,qiu2026qwen38next},
Cosmos-Reason2~\cite{nvidia2026cosmosreason2},
Gemma~4~\cite{gemmateam2026gemma4},
InternVL3.5~\cite{wang2025internvl35},
GLM-4.1V-Thinking~\cite{glmteam2025glm41v},
MiniCPM-V~4.5~\cite{yu2026minicpmv45},
Ovis2.5~\cite{lu2025ovis25},
VideoLLaMA3~\cite{zhang2025videollama3},
and VITA-1.5~\cite{fu2025vita15}. On the real-world set, we additionally evaluate seven
configurations via APIs: the open-weight Kimi K3~\cite{kimiteam2026kimik3}
and six proprietary configurations from
GPT~\cite{openai2026gpt6solluna,openai2026gpt56sol},
Claude~\cite{anthropic2026claudeopus55,anthropic2026claudesonnet55,
anthropic2025claudehaiku45}, and Grok~\cite{xai2026grok47}. %

\vspace{3pt} \noindent
\textbf{Synthetic advantages may reverse in the real world.}
The reversal observed among 8B models in
Section~\ref{sec:motivation} extends to larger configurations.
Gemma~4-31B and InternVL3.5-38B both outperform
Cosmos-Reason2-32B in all four synthetic driving conditions,
yet trail it in all four corresponding real-world conditions.
For Gemma versus Cosmos, the synthetic advantage ranges from
3.4 to 7.0 percentage points when averaging L1--L3 within each
condition; on real-world videos, this becomes a deficit of
5.2--7.4 points.
Reversals across conditions and scales indicate a systematic
shift in relative model strengths.
These comparisons span different source scenes and questions,
so the reversal reflects a broader distribution shift.

\vspace{3pt} \noindent
\textbf{Overall accuracy masks level-specific strengths.}
Across all four synthetic driving conditions,
Qwen3.8-Flash-Next outperforms Qwen3.6-27B at L1 but trails
at both L2 and L3.
A similar reversal appears on the real-world set:
Kimi K3 achieves the highest overall accuracy among the
24 open-weight configurations ($85.4\%$), exceeding
Qwen3.8-Flash-Next ($83.2\%$), yet scores lower at L1
($86.7\%$ vs.\ $89.5\%$).
Its advantage instead lies in L2 and L3, where it achieves
$88.8\%$ and $80.8\%$, compared with $82.5\%$ and $77.6\%$.
A model's overall advantage therefore need not extend to
fundamental visual judgments, motivating their evaluation
separately from more demanding reasoning.

\findingbox{1}{Model rankings can reverse across synthetic and real-world conditions, and overall accuracy can conceal distinct strengths across difficulty levels.}

\vspace{3pt} \noindent
\textbf{Model scaling generally improves accuracy.}
Qwen3.5 improves monotonically across all five evaluated sizes
on both partitions: overall accuracy rises from $56.0\%$ to
$80.3\%$ on synthetic videos and from $65.7\%$ to $82.2\%$
on real-world videos between 0.8B and 27B.
InternVL3.5 provides an exception: its 14B variant improves
over 8B on synthetic videos ($68.8\%$ vs.\ $67.2\%$), but
falls behind on real-world videos ($67.6\%$ vs.\ $68.4\%$).
Across families, the parameter--accuracy trade-off is also uneven.
Kimi K3 uses 2.8T total parameters, $22.4\times$ the 125B of
Qwen3.8-Flash-Next, yet improves real-world accuracy by only
2.2 points ($85.4\%$ vs.\ $83.2\%$).
Thus, substantially larger parameter counts can accompany
modest accuracy gains; size alone does not determine the
benefit under natural degradation.

\vspace{3pt} \noindent
\textbf{Scaling does not consistently reduce corruption losses.}
All 23 configurations evaluated on the synthetic set lose
accuracy relative to their clean-video counterparts, with drops
of 7.8--11.4 percentage points
(Table~\ref{tab:grape_main_levels}).
Even Qwen3.6-27B, which has the smallest gap, loses 7.8 points.
Within Qwen3.5, increasing size from 0.8B to 2B raises
corrupted accuracy from $56.0\%$ to $63.8\%$, while the drop
increases from 10.4 to 11.3 points.
At 27B, accuracy reaches $80.3\%$ and the drop narrows to
7.9 points.
Reporting both measures distinguishes performance under
degradation from the loss relative to clean observations.
Appendix~\ref{app:per-corruption} breaks these losses down by corruption type and difficulty level.

\findingbox{2}{Model scaling generally improves accuracy, but more parameters do not guarantee better real-world performance or smaller corruption-induced losses.}

\vspace{3pt} \noindent
\textbf{Frontier models.}
GPT-5.6 Sol achieves the highest accuracy on the real-world set, while Kimi K3 outperforms four of the six proprietary models evaluated (Appendix~\ref{app:proprietary-comparisons}).

\section{Synthetic-to-Real Transfer}
\label{sec:exp}

The synthetic set provides answer supervision for corrupted inputs and paired clean videos for teacher guidance. We use these resources to test whether training improves accuracy on naturally degraded driving videos. Using Qwen3.5-9B~\cite{qwen2026qwen35}, we compare three recipes with rank-16 LoRA~\cite{hu2022lora} and 32-frame inputs, drawing on a pool of 30,924 synthetic training questions. Synthetic training, validation, and test splits are disjoint by source video; all model selection uses synthetic validation data. The real-world set is used only for testing in this transfer experiment. Appendix~\ref{sec:supp_training} provides split and optimization details.

\subsection{Training Methods}
\label{sec:exp_methods}

\begingroup
\setlength{\abovedisplayskip}{5pt plus 1pt minus 1pt}
\setlength{\belowdisplayskip}{5pt plus 1pt minus 1pt}
\setlength{\abovedisplayshortskip}{1pt}
\setlength{\belowdisplayshortskip}{3pt plus 1pt minus 1pt}

Let $x=(\tilde{v},q,\mathcal{A})$ and $x^0=(v^0,q,\mathcal{A})$ denote the corrupted and clean inputs, respectively, and let $\pi_\theta$ be the student's conditional response distribution.

\vspace{3pt} \noindent
\textbf{Supervised fine-tuning (SFT).}
We train the student for one epoch to predict only the correct option:
\begin{equation}
    \mathcal{L}_{\mathrm{SFT}}=-\mathbb{E}_{x,y}\!\left[\log\pi_\theta(y\mid x)\right].
    \label{eq:sft}
\end{equation}

\vspace{3pt} \noindent
\textbf{Supervised fine-tuning with reasoning.}
The unadapted model generates rationales $r$ from $x^0$, using answer-conditioned rationalization when needed~\cite{zelikman2022star}. We retain those ending in the correct answer $y$ and train for one epoch on $z=(r,y)$:
\begin{equation}
    \mathcal{L}_{\mathrm{RSFT}}=-\mathbb{E}_{x,z}\!\left[\textstyle\sum_t\log\pi_\theta(z_t\mid x,z_{<t})\right].
    \label{eq:rsft}
\end{equation}

\vspace{3pt} \noindent
\textbf{On-policy self-distillation (OPSD).}
Starting from reasoning SFT, we train for one additional epoch~\cite{agarwal2024onpolicy}. The student samples a response $\hat{z}\sim\pi_\theta(\cdot\mid x)$; a frozen copy of the starting model serves as teacher and observes the clean input $x^0$ and reference answer $y$. Both models score the same student-generated prefixes, yielding next-token distributions $p_{\theta,t}$ (student) and $p_{T,t}$ (teacher):
\begin{equation}
    \mathcal{L}_{\mathrm{OPSD}}=\mathbb{E}_{x,\hat{z}}\!\left[\tfrac{1}{|\hat{z}|}\textstyle\sum_t D_t+\lambda\ell_{\mathrm{ans}}\right],
    \label{eq:opsd}
\end{equation}
where $D_t=\mathrm{KL}(p_{\theta,t}\|p_{T,t})$ is reverse KL and $\ell_{\mathrm{ans}}$ is cross-entropy on the reference answer ($\lambda=0.5$). Appendix~\ref{app:transfer-objectives} gives the full conditioning.
\par\endgroup

\subsection{Main Results and Analysis}
\label{sec:exp_results}

\begin{figure}[t]
    \centering
    \includegraphics[width=\columnwidth]{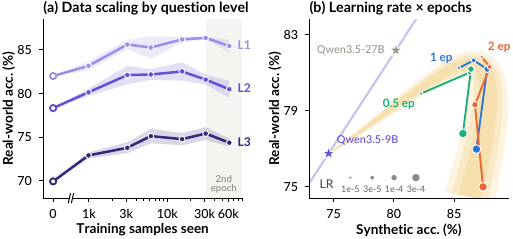}
    \vspace{-12pt}
    \caption{\textbf{Can synthetic training transfer to real-world scenarios?} (a) Real-world accuracy at each question level versus the number of synthetic training questions seen. Curves and bands show the mean and standard deviation across three seeds; gray shading marks the second training epoch. (b) Learning rate and epoch sweeps, evaluated on synthetic validation and real-world data. Colors encode epochs and marker sizes encode learning rates. Hollow markers and stars denote unadapted models. All SFT runs use Qwen3.5-9B; (a) fixes the learning rate at $10^{-4}$.}
    \label{fig:synthetic-to-real-transfer}
    \vspace{-5pt}
\end{figure}

\noindent
\textbf{Benefits of synthetic training.}
Table~\ref{tab:grape_realworld_levels} shows improvements over the unadapted Qwen3.5-9B baseline: direct SFT raises real-world accuracy from 76.7\% to 81.1\%, improving all four conditions and all three question levels. L3 gains most, rising from 69.9\% to 75.4\%. SFT with reasoning reaches 78.1\% despite its longer training targets. One possible explanation is that direct supervision of answer correctness is sparse within these longer traces. OPSD adds guidance from the clean-video teacher's token distributions along the student's own reasoning trajectories, raising accuracy to 80.4\% and approaching direct SFT. %

\vspace{3pt} \noindent
\textbf{Data scaling.}
Figure~\ref{fig:synthetic-to-real-transfer}(a) shows how transfer changes as more synthetic training questions are seen. L2 rises from 78.3\% to around 82\% within the first 3K questions, with little further gain from larger budgets. L3 benefits over a wider range of training sizes, rising from 69.9\% to 75.4\% after one full pass through the pool. L1 similarly obtains most of its gains early. A second pass over the same questions lowers mean accuracy at all three levels, showing diminishing returns from additional training exposure.

\vspace{3pt} \noindent
\textbf{Learning rate and epoch sweep.}
Figure~\ref{fig:synthetic-to-real-transfer}(b) shows that higher learning rates and longer training can hurt transfer. At learning rate $3{\times}10^{-4}$, increasing training from 0.5 to 2 epochs raises synthetic validation accuracy from 85.7\% to 87.4\% but lowers real-world accuracy from 77.8\% to 75.0\%. With the same two-epoch budget, a smaller learning rate of $10^{-5}$ reaches 81.8\% real-world accuracy despite nearly identical synthetic accuracy (87.3\%). These trends are consistent with overfitting to synthetic conditions: stronger fitting within the synthetic domain can come at the expense of real-world generalization.

\findingbox{3}{Synthetic SFT and clean-privileged on-policy distillation improve real-world accuracy, although fitting synthetic data more closely can hurt transfer.}

\section{Preference Steering across Difficulty Levels}
\label{sec:method}

\grapebench{} motivates balancing safety-relevant basic judgments with demanding reasoning. We test whether a frozen VLM can redistribute accuracy across these difficulty levels while keeping overall accuracy stable. Inspired by Bi-directional Preference Optimization (BiPO)~\cite{cao2024bipo}, we learn two vectors for continuous preference control. \looseness=-1

\subsection{Preference-Conditioned Steering}

\begingroup
\setlength{\abovedisplayskip}{5pt plus 1pt minus 1pt}
\setlength{\belowdisplayskip}{5pt plus 1pt minus 1pt}
\setlength{\abovedisplayshortskip}{1pt}
\setlength{\belowdisplayshortskip}{3pt plus 1pt minus 1pt}

\noindent\textbf{Steering vectors.}
We learn $\mathbf{u},\mathbf{d}\in\mathbb{R}^{D}$ and set
\begin{equation}
    \mathbf{s}(\lambda)=\mathbf{u}+\lambda\mathbf{d},
    \qquad \lambda\in[-1,1].
    \label{eq:knob}
\end{equation}
The shared component $\mathbf{u}$ is retained across settings, while $\mathbf{d}$ controls the relative emphasis on L1 and L3. The endpoints $\lambda=-1$ and $\lambda=+1$ specify L3 and L1 preferences, respectively, with intermediate values interpolating between them. At decoder layer $\ell$, we apply the intervention to each textual position $t$ following the video tokens:
\begin{equation}
    \mathbf{h}_{\ell,t}\leftarrow\mathbf{h}_{\ell,t}+\mathbf{s}(\lambda).
\end{equation}
The backbone remains frozen, so only $2D$ parameters are learned. At inference, varying $\lambda$ adjusts the intervention without retraining. Setting $\lambda=0$ still applies $\mathbf{u}$; the unsteered reference uses no intervention.

\vspace{3pt} \noindent
\textbf{Preference objective.}
We combine a DPO-style loss~\cite{rafailov2023dpo} with preference-dependent weighting of the two level objectives, drawing on multi-objective alignment~\cite{zhou2024modpo,zhong2024panacea}. For each video-question prompt $x$, we pair the correct answer $y^+$ with the highest-scoring incorrect option $y^-$ under the unsteered reference $\pi_{\mathrm{ref}}$. Let $\pi_\lambda$ denote the model with intervention $\mathbf{s}(\lambda)$. Its relative preference margin is
\begin{equation}
    m_\lambda(x)=\log\frac{\pi_\lambda(y^+\mid x)}{\pi_\lambda(y^-\mid x)}-\log\frac{\pi_{\mathrm{ref}}(y^+\mid x)}{\pi_{\mathrm{ref}}(y^-\mid x)}.
    \label{eq:preference_margin}
\end{equation}
For $k\in\{1,3\}$, we define the level-specific loss
\begin{equation}
    \mathcal{L}_k(\lambda)=-\mathbb{E}_{x\sim\mathcal{D}_k}\!\left[\log\sigma\big(\beta m_\lambda(x)\big)\right],
\end{equation}
where $\mathcal{D}_k$ contains level-$k$ training questions, $\sigma$ is the sigmoid function, and $\beta>0$ scales the margin. Sampling $\lambda\sim\mathcal{U}[-1,1]$ at each step gives the weighted objective
\begin{equation}
    \mathcal{L}=\mathbb{E}_\lambda\!\left[\frac{1+\lambda}{2}\mathcal{L}_1(\lambda)+\frac{1-\lambda}{2}\mathcal{L}_3(\lambda)\right].
    \label{eq:knobloss}
\end{equation}
Both losses favor correct answers, with $\lambda$ adjusting their relative weights. Overall accuracy is not constrained by the loss and is assessed empirically.
\par\endgroup

\subsection{Evaluating Preference Control}

\noindent\textbf{Experimental setup.}
We use Qwen3.5-9B with 32 sampled video frames and single-letter answer predictions. Two 4096-dimensional vectors are injected at decoder layer $\ell=12$ of 32, giving 8,192 trainable parameters. We learn a separate vector pair for each of the synthetic and real-world driving domains.  Training uses L1 and L3 questions, while overall accuracy pools held-out questions across L1--L3. 

\begin{figure}[t]
    \centering
    \includegraphics[width=\columnwidth]{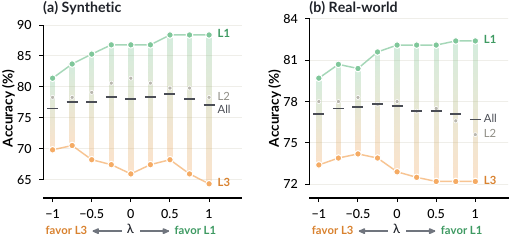}
    \vspace{-15pt}
    \caption{\textbf{Steering accuracy across difficulty levels.} Held-out Qwen3.5-9B accuracy on (a) synthetic and (b) real-world driving videos, using vectors trained separately in each domain. Green and orange markers show L1 and L3; vertical bands span their accuracy gap. Gray dots show L2, and dark horizontal ticks show overall accuracy (All, pooling L1--L3). Increasing $\lambda$ favors L1 over L3; $\lambda=0$ retains the shared vector $\mathbf{u}$ and is not the unsteered model. The panels use different vertical scales.}
    \label{fig:perception-reasoning-steering}
    \vspace{-5pt}
\end{figure}

\vspace{3pt} \noindent
\textbf{Results.}
Figure~\ref{fig:perception-reasoning-steering} shows that steering changes how accuracy is distributed across question levels. On synthetic driving videos, moving from $\lambda=-1$ to $+1$ raises L1 accuracy from $81.4\%$ to $88.4\%$ and lowers L3 accuracy from $69.8\%$ to $64.3\%$, widening the gap between basic judgments and harder reasoning. Overall accuracy varies within $76.5\%$--$78.8\%$, a smaller range than either endpoint shift.
The real-world experiment exhibits the same trade-off: L1 rises from $79.7\%$ to $82.4\%$, while L3 falls from $73.4\%$ to $72.2\%$. Overall accuracy stays within $76.7\%$--$77.8\%$. The endpoint trade-off is consistent across domains, although intermediate settings do not yield strictly monotonic changes. \looseness=-1

\findingbox{4}{Preference steering shifts the balance between safety-relevant judgments and demanding reasoning with little change in overall accuracy.}

\section{Related Work}
\label{sec:related}

\noindent\textbf{Video Understanding Benchmarks.}
Video understanding benchmarks have progressed from broad capability assessment to structured reasoning evaluation~\cite{li2024mvbench,fu2025videomme,nagrani2025minerva}.
NExT-QA and STAR examine causal, temporal, and situated reasoning~\cite{xiao2021nextqa,wu2021star}, while broader suites cover more tasks, domains, and longer videos~\cite{li2024mvbench,fu2025videomme,mangalam2023egoschema,wu2024longvideobench,zhou2025mlvu}.
Perception Test, TempCompass, MotionBench, and TOMATO probe specific perceptual, motion, and temporal skills~\cite{patraucean2023perceptiontest,liu2024tempcompass,hong2025motionbench,shangguan2025tomato}.
VCR-Bench and MINERVA analyze annotated reasoning traces~\cite{qi2025vcrbench,nagrani2025minerva}, while VideoReasonBench probes increasingly difficult tasks involving observations, latent states, and future predictions~\cite{liu2026videoreasonbench}.
More recently, Video-MME-v2 introduces capability hierarchies to assess consistency across related queries and reasoning sequences~\cite{fu2026videommev2}.
\grapebench{} extends these evaluations to visual degradation, examining performance across question levels under synthetic and natural adverse conditions.

\vspace{3pt} \noindent
\textbf{Robustness of Multimodal Models.}
Multimodal robustness depends on visual input quality and reasoning over degraded evidence~\cite{qiu2025mllmic,tang2026robustr1,he2026rova}.
At the image level, R-Bench studies acquisition and processing distortions~\cite{li2025rbench}, while MLLM-IC evaluates corruption types and severity levels~\cite{qiu2025mllmic}.
Video benchmarks examine spatial and temporal tampering in MVTamperBench~\cite{agarwal2025mvtamperbench}, and answer accuracy and reasoning under synthetic disturbances in PVRBench~\cite{he2026rova}.
On the method side, Robust-R1 employs degradation-aware reasoning and reward-driven alignment~\cite{tang2026robustr1}, while Robust-U1 combines visual self-recovery with reasoning over corrupted and recovered images~\cite{tang2026robustu1}. More recently, ROVA encourages consistency between clean and corrupted videos~\cite{he2026rova}.
\grapebench{} complements these efforts with graded questions across synthetic corruptions and naturally adverse conditions, probing video reasoning robustness at different difficulty levels. \looseness=-1

\vspace{3pt} \noindent
\textbf{Driving Benchmarks under Adverse Conditions.}
Driving benchmarks connect visual understanding with prediction and decision-making.
LingoQA and nuScenes-QA evaluate driving scene understanding through QA~\cite{marcu2024lingoqa,qian2024nuscenesqa}, while DriveLM organizes perception, prediction, and planning questions as a graph~\cite{sima2024drivelm}.
DriveXQA uses multiple sensor modalities for question answering in adverse driving scenes~\cite{tao2026drivexqa}.
DriveBench evaluates driving VLMs under clean, corrupted, and text-only inputs, revealing limitations in visual grounding~\cite{xie2025drivebench}.
ObsDriveBench studies naturally adverse weather using camera, LiDAR, and radar observations, assessing observability awareness, spatial reliability, and risk-aware decisions~\cite{yan2026obsdrivebench}.
PaSBench-Video evaluates proactive safety warnings in streaming video across several domains, including driving~\cite{zhao2026pasbenchvideo}.
Unlike these capability-oriented decompositions, \grapebench{} introduces a human-annotated real-world driving-video set with three difficulty-graded questions per video.
Combined with its synthetic daily-life and driving sets, it enables robustness evaluation across visual conditions and reasoning demands while distinguishing difficulty from safety relevance.

\section{Conclusion}
\label{sec:conclusion}

We introduced \grapebench{} to study video reasoning under synthetic and natural visual degradation through three questions of increasing reasoning demand per video. Model rankings can reverse between simulated and real-world driving conditions, yet training on the synthetic set improves Qwen3.5-9B's accuracy on real-world footage. We observe this transfer with supervised fine-tuning and, when reasoning is enabled, with on-policy self-distillation from a teacher given access to the clean video. The question tiers reveal that a model with higher overall accuracy can still be less reliable on basic judgments that matter for safe driving. Using two learned steering vectors, we can adjust a frozen model's balance of accuracy between these judgments and more demanding questions, with little change in overall accuracy. These findings highlight the value of testing even simple judgments under naturally adverse conditions when a driving decision depends on them. \looseness=-1

\section*{Acknowledgments}
This work has been funded in part by the ARL award W911QX-24-F-0049, DARPA award FA8750-23-2-1015, ONR award N00014-23-1-2840, ONR MURI grant N00014-25-1-2116, and USDA-NIFA award 2021-67021-35329.

\par
{
    \small
    \bibliographystyle{ieeenat_fullname}
    \bibliography{main}
}

\clearpage\clearpage
\clearpage
\hypersetup{pageanchor=false}
\backrefsetup{disable}
\setcounter{page}{1}
\appendix
\maketitlesupplementary

\section*{Overview}
\begingroup
\setlength{\parskip}{0pt}
In this supplementary material, we provide additional details on benchmark construction, evaluation, and synthetic-to-real training, expand the related-work discussion and our contributions, and discuss limitations and societal impact. The material is organized as follows:
\begin{itemize}[topsep=3pt,itemsep=3pt,partopsep=0pt]
    \item \textbf{Section~\ref{app:synthetic-construction}} describes synthetic question verification and corruption generation.
    \item \textbf{Section~\ref{app:realworld-construction}} covers real-world clip preparation, question annotation, and metadata.
    \item \textbf{Section~\ref{app:evaluation-details}} presents evaluation details, paired corruption analyses, and additional model comparisons.
    \item \textbf{Section~\ref{sec:supp_training}} provides training recipes, distillation objectives, and additional transfer analyses.
    \item \textbf{Section~\ref{app:extended-related}} reviews additional related work and explains our distinct contributions.
    \item \textbf{Section~\ref{app:limitations-impact}} discusses limitations, potential benefits, and risks of downstream use.
\end{itemize}
\endgroup

\section{Synthetic Set Construction}
\label{app:synthetic-construction}

\subsection{Automatic Question Verification}
\label{app:verification}

\noindent
\textbf{Shared verification gates.}
The pipeline implements six shared gates.
\textbf{G0 (option sanity)} checks the option count, nonempty and nonduplicate option text, and consistency between the answer label and the available options.
\textbf{G1 (logic and hierarchy)} checks that questions are self-contained, follow a common visual thread, and satisfy the intended hierarchy.
\textbf{G2 (diversity)} checks for near-paraphrases across levels and, under the embedding-based configuration, repeated L1 questions within the same subcategory.
\textbf{G3 (visual grounding and visual-only wording)} comprises two checks: G3a screens references to narration, audio, or external subtitles, while G3b checks the stated answers against sampled video frames without a transcript.
\textbf{G4 (video necessity)} tests whether a question can be answered from prior knowledge alone, using a text-only response produced without the video or answer options.
\textbf{G5 (distractor plausibility)} checks that options are of the same semantic kind and remain plausible without viewing the video.

\vspace{3pt} \noindent
\textbf{Execution and difficulty audit.}
G1 and G3a precede distractor generation. The subsequent checks run in the order G0, G3b, G4, G5, and G2. An additional difficulty audit (GD) is configured before G2 for EgoSchema, NExT-QA, STAR, and VRU-Accident. It assesses the intended difficulty progression from the final questions, options, and frames without construction metadata or reference answers. 

\vspace{3pt} \noindent
\textbf{Revision and inclusion.}
When a check flags malformed answer options, inconsistent option formats,
or weak distractors, the pipeline regenerates the options at the affected
question levels and reruns verification. If the group still fails, or if
the failure requires changes to the questions, the pipeline re-extracts
visual facts from the sampled frames and uses them to regenerate the full
L1--L3 question group. It reuses the original frames by default and samples
new frames when the verifier requests more visual evidence. As reported in
Section~\ref{sec:data-statistics}, 10,818 groups passed automatic verification.
We manually reviewed, revised, and approved the remaining 23 groups for release.
The metadata record the original automatic verification outcome.

\subsection{Synthetic Corruptions}
\label{app:synthetic-corruptions}

\noindent
\textbf{Corruption set and assignment.}
We use twelve corruption types to cover adverse weather, lens and glass
degradation, exposure changes, blur, and compression or transmission artifacts.
Rain, snow, nighttime, and fog are assigned to the 3,000 question groups from
VRU-Accident. The remaining eight types, including wet lens, glass occlusion,
overexposure, underexposure, defocus blur, motion blur, low-bitrate compression,
and bitstream errors, are assigned to the 7,841 groups from NExT-QA, STAR,
EgoSchema, and Video-MME short/medium. The benchmark uses one fixed parameter configuration
for each corruption type.

Each question group receives one corrupted video shared by its three question
levels, while its questions, answer options, and reference answers remain
unchanged. Assignments are balanced within each source-dataset pool, and
groups derived from the same source video receive different corruption types.
The weather and general-corruption subsets therefore cover different source
content rather than a fully crossed video-by-corruption design.

\vspace{3pt} \noindent
\textbf{Corruption implementation.}
Rain, snow, and nighttime are generated with a procedural renderer operating
in linear-light space. Rain and snow use multiple particle layers with
temporally coherent trajectories, combined with illumination changes and
haze. Rain additionally approximates wet-road reflections, whereas snow
represents airborne flakes rather than accumulated ground snow. Nighttime
combines reduced ambient illumination with approximate headlights, street
lighting, glare, and signal-dependent noise. These effects use heuristic
scene assumptions rather than reconstructed scene geometry. Fog is generated
separately using smoothly varying, drifting opacity fields and contrast
attenuation.

Wet-lens corruption adapts the droplet-refraction effect provided in
HiKER-SGG~\cite{zhang2024hiker}. Fixed, multi-scale droplet fields spatially
remap each input frame, producing persistent lens-attached distortions rather
than overlaying unrelated scene content. Glass occlusion combines
soft-edged translucent masks, local blur, and tint to approximate obscured
glass. Defocus uses Gaussian blur, while motion blur integrates neighboring
frames under simulated camera shake to produce temporal trails.
Overexposure and underexposure use intensity scaling and nonlinear tone
mapping, with clipping in the overexposed case.

Low-bitrate corruption uses constrained H.264 encoding through FFmpeg.
Following prior work on video corruption
robustness~\cite{yi2021minikineticsc}, we introduce bitstream errors
directly into encoded video data rather than drawing image-space blocks. The damaged
stream is decoded and re-encoded into a valid video, preserving visible
decoding artifacts for consistent playback. 
We will release the code so readers can inspect how each corruption is applied and which parameters it uses.

\section{Real-World Set Construction}
\label{app:realworld-construction}

\subsection{Video Sources and Preparation}

We collect real-world driving footage from
DrivingDojo~\cite{wang2024drivingdojo},
Boreas~\cite{burnett2023boreas},
Boreas-RT~\cite{lisus2026boreasrt},
BDD100K~\cite{yu2020bdd100k}, and
DrivingHazy/GoProHazy~\cite{fan2024dvd}.
Candidate clips are identified using available weather and
time-of-day metadata and screened for suitability for
question annotation.

We reconstruct DrivingDojo videos from the provided image
sequences, extract 15-second clips from continuous Boreas
and Boreas-RT recordings, and retain up to the first
30 seconds of each BDD100K and DrivingHazy/GoProHazy video.
Shorter videos retain their full temporal extent.
All clips are prepared at 5 fps without adding synthetic
weather effects.
DrivingDojo clips retain their source resolution of
$1920\times1080$, while clips from the other sources are
prepared at a width of 1280 pixels with their aspect ratios
preserved.
Real-world atmospheric haze in DrivingHazy and GoProHazy
is included in the fog category for analysis.

\subsection{Question Design and Annotation}
\label{app:realworld-annotation}

Expert annotators developed the real-world questions, writing some directly and using
AI-generated drafts as a starting point for others. They inspected each clip using
playback and frame stepping and excluded unsuitable clips. For each retained clip,
the annotators finalized three self-contained questions, one per difficulty level,
after checking each question, its four options, and its reference answer against
the video and revising or replacing questions as needed.
Looking across the completed real-world set, we observed the following broad patterns
in what questions at each difficulty level ask:
\begin{itemize}
    \item \textbf{L1} questions often ask about a salient entity, event, or scene state,
    such as whether a pedestrian is present or a nearby vehicle is moving.

    \item \textbf{L2} questions often concern a specific attribute, spatial relation,
    or change over time, such as how another vehicle moves relative to the ego vehicle
    or how the lane layout changes during a maneuver.

    \item \textbf{L3} questions often combine several observations, for example
    asking how many vehicles passed and in what order.
\end{itemize}

\begin{figure*}[t]
    \centering
    \includegraphics[width=\textwidth]{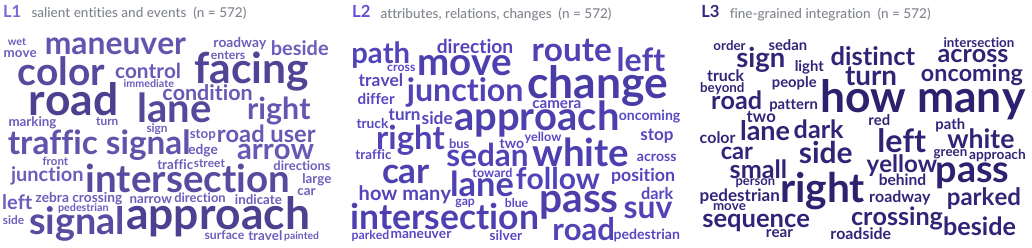}
    \caption{\textbf{Characteristic words of the real-world questions at each level.}
    One word cloud per level (572 questions each). Word size reflects how characteristic a word is of that level:
    its count in the level, weighted by its relative frequency in that level versus all 1,716 questions, so that words
    shared by every level, such as ``ego vehicle'' and ``video'', do not dominate; stop words and generic question phrasing are removed.
    L1 questions center on salient scene elements such as signals, lanes, and intersections; L2 shifts toward motion and
    relations, with words such as change, approach, follow, and route; and L3 is dominated by counting (``how many''), left/right
    distinctions, and sequences of passing objects.}
    \label{fig:question-wordcloud}
\end{figure*}

Questions share the visual context but are answered
independently, without reference answers from other levels.
Figure~\ref{fig:question-wordcloud} shows the words that characterize each level.

\vspace{3pt} \noindent
\textbf{Temporal annotations.}
A prepared driving clip may contain multiple events and
road users.
We therefore allow annotators to specify a question-specific
observation endpoint when constructing questions and
reference answers.
This allows annotators to control the amount and temporal
scope of evidence relevant to each question, helping align
its evidence demands with the intended question level.
For example, a shorter prefix can reduce ambiguity about
the target entity or event in an L1 question, while a
specified observation endpoint can support prediction
about subsequent events.
The \texttt{cutoff\_sec} field defines this endpoint:
a cutoff of $t$ seconds restricts the model's input to
the first $t$ seconds of the prepared clip.
If no cutoff is specified, the full clip is used.

\vspace{3pt} \noindent
\textbf{Option balancing.}
To reduce answer-position bias, options are permuted and
answer letters are updated while preserving the correct
option content.
Across the 1,716 questions, A, B, C, and D each occur
429 times as the correct answer.
Balancing additionally covers condition, level, source
dataset, question family, operation, and selected
intersections.

\subsection{Condition and Question-Type Metadata}

\noindent
\textbf{Condition labels.}
For each clip, \texttt{source\_condition} records the condition label
from the source dataset, and \texttt{observed\_condition} records the
condition identified by the annotator. For evaluation, we use the
standardized label in \texttt{condition}, which assigns each clip to
one of five categories: \emph{rain}, \emph{snow}, \emph{night},
\emph{fog}, or \emph{others}. We group haze under \emph{fog} and assign
conditions outside the four named categories to \emph{others}.

\vspace{3pt} \noindent
\textbf{Question types.}
Each question has two type labels in addition to its L1--L3
level. The \texttt{question\_family} field describes the broad
task: \emph{perception} concerns visible entities, states,
or events; \emph{reasoning} concerns explanations or evidence
integration; \emph{planning} concerns actions or maneuvers;
and \emph{prediction} concerns future states or intentions.
The \texttt{operation\_tag} field identifies a more specific
task, such as counting, spatial relations, or causal reasoning.

Question-type distributions vary across levels, but the
hierarchy does not impose a fixed mapping between type and
level.
As shown in Table~\ref{tab:realworld-operations},
perception accounts for 488 of the 572 L1 questions,
while 143 of the 160 multi-fact inference questions occur
at L3.
Nevertheless, all four question families appear at every
level, and L3 includes 188 perception questions.
Question type therefore complements the level annotation
by describing the task rather than determining its difficulty.

\begin{table}[t]
\centering
\small
\setlength{\tabcolsep}{4pt}
\begin{tabular}{lrrrr}
\toprule
\textbf{Question type}
& \textbf{L1}
& \textbf{L2}
& \textbf{L3}
& \textbf{Total} \\
\midrule
\multicolumn{5}{l}{\textit{Question family}} \\
Perception
& 488 & 134 & 188 & 810 \\
Reasoning
& 52 & 426 & 315 & 793 \\
Planning
& 26 & 9 & 45 & 80 \\
Prediction
& 6 & 3 & 24 & 33 \\
\midrule
\multicolumn{5}{l}{\textit{Operation}} \\
Attribute / state
& 165 & 21 & 95 & 281 \\
Trajectory / motion
& 65 & 174 & 25 & 264 \\
Spatial relation
& 71 & 102 & 19 & 192 \\
Counting
& 24 & 58 & 94 & 176 \\
Entity recognition
& 118 & 18 & 33 & 169 \\
Multi-fact inference
& 7 & 10 & 143 & 160 \\
Temporal relation
& 6 & 73 & 38 & 117 \\
Causal reasoning
& 24 & 35 & 21 & 80 \\
Hazard / risk assessment
& 14 & 23 & 33 & 70 \\
Action / event
& 41 & 11 & 7 & 59 \\
Ego-action planning
& 14 & 14 & 23 & 51 \\
Route / maneuver planning
& 10 & 15 & 10 & 35 \\
Intent prediction
& 7 & 7 & 19 & 33 \\
Right-of-way reasoning
& 5 & 11 & 6 & 22 \\
Future-state prediction
& 1 & 0 & 6 & 7 \\
\midrule
\textbf{Total questions}
& \textbf{572} & \textbf{572} & \textbf{572}
& \textbf{1,716} \\
\bottomrule
\end{tabular}
\caption{Question-type distributions across levels in the
real-world set. Each question has one family label and one
operation label; both sections describe the same 1,716
questions.}
\label{tab:realworld-operations}
\end{table}

\section{Evaluation Details and Additional Analyses}
\label{app:evaluation-details}

\subsection{Evaluation Protocol}
\label{app:evaluation-protocol}

\noindent
\textbf{Evaluated samples.}
The synthetic evaluation uses all 10,841 question groups
(32,523 questions).
Clean and corrupted evaluations use identical question IDs,
questions, options, and reference answers.
The real-world evaluation covers all 572 videos and 1,716
questions, including the 20 videos labeled as \emph{others}.
Every reported baseline configuration has one scored
prediction per evaluated question.
Accuracy is pooled over questions, and the synthetic
$\Delta$ is the paired difference between corrupted and
clean accuracy in percentage points.

\vspace{3pt} \noindent
\textbf{Video inputs.}
Synthetic questions use the full clip.
For real-world questions, video-input configurations receive
a losslessly re-encoded prefix ending at the annotated
cutoff, preserving the source resolution and frame rate.
Frames are then sampled by the model-specific video loader
(Table~\ref{tab:evaluation-settings}).
API models and Ovis2.5 instead receive pre-extracted images:
up to 32 frames sampled at bin-center timestamps within the
permitted prefix, resized to at most 768 pixels on the
long side and JPEG-encoded at quality 75.
For synthetic Ovis2.5 evaluations, up to 32 frames are
extracted at a uniform rate over the full clip.
No audio, subtitles, or reference annotations are supplied.
API inputs contain chronologically ordered images without
timestamps; native video processors retain their own
timestamp-handling conventions.

\begin{table}[t]
\centering
\small
\setlength{\tabcolsep}{3pt}
\renewcommand{\arraystretch}{1.1}
\begin{tabular}{@{}p{0.32\linewidth}p{0.48\linewidth}r@{}}
\toprule
\textbf{Model group} &
\textbf{Frame sampling} &
\textbf{Tokens} \\
\midrule
Qwen &
Target 2 fps; 4--768 frames &
1,024 \\
Cosmos-Reason2 &
Target 2 fps; 4--768 frames &
4,096 \\
Gemma~4 &
32 uniform frames; all if fewer &
1,024 \\
InternVL3.5 &
32 uniform frames; all if fewer &
1,024 \\
MiniCPM-V~4.5 &
32 uniform frames; all if fewer &
1,024 \\
GLM-4.1V-Thinking &
2 fps up to 300 s; otherwise 600 uniform frames &
8,192 \\
Ovis2.5 &
Up to 32 extracted images &
1,024 \\
VideoLLaMA3 &
1 fps; at most 64 frames &
16 \\
VITA-1.5 &
1 fps; at most 16 frames, padded to at least 4 &
16 \\
Claude &
Up to 32 extracted images &
1,000 \\
GPT &
Up to 32 extracted images &
4,000 \\
Grok, Kimi &
Up to 32 extracted images &
8,000 \\
\bottomrule
\end{tabular}
\caption{Model-specific sampling policies and maximum output
token budgets for baseline evaluations.
Frame counts depend on the available clip length.
Qwen and Cosmos use a total video pixel budget of
25,165,824; InternVL uses one $448\times448$ tile per frame,
and MiniCPM uses one 448-pixel slice per frame.
The token limits include reasoning where applicable.}
\label{tab:evaluation-settings}
\end{table}

\vspace{3pt} \noindent
\textbf{Runtime and precision.}
Most locally evaluated models run in BF16 with vLLM 0.31.0. We use official FP8 checkpoints for Flash-Next,
Qwen3.8-27B, and Qwen3.6-27B, and online FP8 weight quantization
for InternVL3.5-38B. FP8 KV caches are enabled for Qwen3.5-27B,
Qwen3.6-27B, and Qwen3.8-27B. VideoLLaMA3 runs in BF16 with
Transformers 4.46.3, while VITA-1.5 runs in FP16 with
Transformers 4.43.4.

\vspace{3pt} \noindent
\textbf{Prompting and decoding.}
We evaluate the L1--L3 questions for each video separately.
The model receives one question at a time, together with its video
input and answer options. Answers and rationales from other levels
are not provided.
We format inputs using model-specific chat templates.
Baseline prompts instruct the model to return only the correct
option letter.
Locally evaluated models use greedy decoding.
We explicitly disable thinking for Qwen3.5/3.6/3.8 and Flash-Next.
GLM-4.1V-Thinking and Ovis2.5 retain their reasoning behavior.
API requests use provider-default sampling without an explicit
temperature. For reasoning, we use low effort for Claude Opus
and Sonnet and provider defaults for GPT, Grok, and Kimi.

\vspace{3pt} \noindent
\textbf{Answer extraction and failures.}
We read the option letter (A--D) from the expected answer format, such as ``B'' or ``Answer: B''.
If the format differs, we scan the answer text for an option letter. A response with no extractable option counts as incorrect.
We retry requests only after infrastructure failures.

\subsection{Per-Corruption Accuracy Changes}
\label{app:per-corruption}

Table~\ref{tab:per_corruption_delta} summarizes the paired clean-to-corrupted accuracy change for each corruption type, averaged over the 23 open models evaluated on the synthetic set, and splits it by question level. Table~\ref{tab:per_model_corruption_delta} lists the change for every model and corruption. Each comparison uses the same questions with the clean source video and with its corrupted version, so the changes measure losses relative to each subset's own clean baseline; driving and daily-life corruptions apply to different source videos.

\begin{table}[t]
\centering
\caption{\textbf{Per-corruption accuracy changes on the synthetic set.} Paired accuracy (\%) on the clean source videos and on their corrupted versions for the same questions, averaged over the 23 open models of Table~\ref{tab:grape_main_levels}. $\Delta$ is corrupted minus clean (percentage points); the range spans the 23 models, and the last three columns split the mean $\Delta$ by level.}
\label{tab:per_corruption_delta}
\setlength{\tabcolsep}{3pt}
\resizebox{\linewidth}{!}{%
\begin{tabular}{llrrrrcrrr}
\toprule
 & Corruption & \#Q & Clean & Corr. & $\Delta$ & $\Delta$ range & L1 & L2 & L3 \\
\midrule
\multirow{4}{*}{\rotatebox{90}{Driving}} & Rain & 2,250 & $72.0$ & $63.2$ & ${-}8.9$ & [${-}12.0$, ${-}6.1$] & ${-}12.2$ & ${-}7.5$ & ${-}7.0$ \\
 & Snow & 2,250 & $70.9$ & $61.4$ & ${-}9.5$ & [${-}11.9$, ${-}6.4$] & ${-}13.9$ & ${-}7.7$ & ${-}6.7$ \\
 & Night & 2,250 & $72.1$ & $65.7$ & ${-}6.4$ & [${-}9.4$, ${-}4.5$] & ${-}7.5$ & ${-}6.1$ & ${-}5.7$ \\
 & Fog & 2,250 & $72.3$ & $63.6$ & ${-}8.7$ & [${-}12.6$, ${-}5.0$] & ${-}11.4$ & ${-}7.4$ & ${-}7.2$ \\
\midrule
\multirow{8}{*}{\rotatebox{90}{Daily}} & Wet lens & 2,943 & $82.1$ & $71.5$ & ${-}10.6$ & [${-}12.7$, ${-}7.7$] & ${-}12.3$ & ${-}11.2$ & ${-}8.3$ \\
 & Glass & 2,943 & $80.8$ & $60.6$ & ${-}20.1$ & [${-}24.2$, ${-}15.3$] & ${-}27.1$ & ${-}19.3$ & ${-}14.0$ \\
 & Overexp. & 2,943 & $81.6$ & $75.9$ & ${-}5.7$ & [${-}7.0$, ${-}4.3$] & ${-}6.9$ & ${-}5.5$ & ${-}4.7$ \\
 & Underexp. & 2,934 & $80.7$ & $71.1$ & ${-}9.6$ & [${-}14.8$, ${-}6.5$] & ${-}11.8$ & ${-}9.7$ & ${-}7.4$ \\
 & Defocus & 2,943 & $81.6$ & $67.5$ & ${-}14.1$ & [${-}19.4$, ${-}10.8$] & ${-}17.5$ & ${-}14.3$ & ${-}10.6$ \\
 & Motion & 2,940 & $80.6$ & $73.5$ & ${-}7.1$ & [${-}9.9$, ${-}4.5$] & ${-}7.8$ & ${-}7.9$ & ${-}5.6$ \\
 & Bitrate & 2,943 & $80.6$ & $69.6$ & ${-}11.0$ & [${-}14.7$, ${-}8.0$] & ${-}13.6$ & ${-}11.0$ & ${-}8.5$ \\
 & Bit error & 2,934 & $80.7$ & $76.4$ & ${-}4.4$ & [${-}6.4$, ${-}2.9$] & ${-}4.9$ & ${-}4.7$ & ${-}3.6$ \\
\bottomrule
\end{tabular}}
\end{table}

\begin{table*}[t]
\centering
\caption{\textbf{Accuracy change per model and corruption on the synthetic set.} Corrupted minus clean accuracy (percentage points) on the same questions. Avg.\ averages the twelve corruptions equally; Table~\ref{tab:grape_main_levels} instead reports the question-weighted change over all questions.}
\label{tab:per_model_corruption_delta}
\setlength{\tabcolsep}{3.2pt}
\resizebox{\textwidth}{!}{%
\begin{tabular}{llrrrrrrrrrrrrr}
\toprule
 & & \multicolumn{4}{c}{Driving} & \multicolumn{8}{c}{Daily} & \\ \cmidrule(lr){3-6} \cmidrule(lr){7-14}
Model & Size & Rain & Snow & Night & Fog & Wet lens & Glass & Overexp. & Underexp. & Defocus & Motion & Bitrate & Bit error & Avg. \\
\midrule
Qwen3.8-Flash-Next & 125B & ${-}8.8$ & ${-}11.9$ & ${-}7.6$ & ${-}8.0$ & ${-}9.3$ & ${-}19.8$ & ${-}5.2$ & ${-}8.0$ & ${-}12.0$ & ${-}4.8$ & ${-}8.1$ & ${-}4.5$ & ${-}9.0$ \\
Qwen3.8 & 27B & ${-}9.4$ & ${-}9.0$ & ${-}8.7$ & ${-}9.2$ & ${-}10.4$ & ${-}16.4$ & ${-}5.7$ & ${-}9.1$ & ${-}12.6$ & ${-}5.4$ & ${-}9.4$ & ${-}4.2$ & ${-}9.1$ \\
Qwen3.6 & 27B & ${-}9.2$ & ${-}9.4$ & ${-}5.0$ & ${-}8.4$ & ${-}7.9$ & ${-}15.3$ & ${-}4.8$ & ${-}7.2$ & ${-}11.1$ & ${-}4.7$ & ${-}8.2$ & ${-}2.9$ & ${-}7.8$ \\
Qwen3.5 & 27B & ${-}8.8$ & ${-}9.6$ & ${-}5.5$ & ${-}8.1$ & ${-}8.5$ & ${-}16.4$ & ${-}4.7$ & ${-}7.4$ & ${-}10.8$ & ${-}4.5$ & ${-}8.0$ & ${-}3.2$ & ${-}7.9$ \\
Qwen3.5 & 9B & ${-}8.5$ & ${-}10.3$ & ${-}6.5$ & ${-}8.7$ & ${-}9.8$ & ${-}19.3$ & ${-}5.1$ & ${-}9.6$ & ${-}14.4$ & ${-}5.4$ & ${-}9.3$ & ${-}4.7$ & ${-}9.3$ \\
Qwen3.5 & 4B & ${-}9.4$ & ${-}11.2$ & ${-}6.9$ & ${-}10.5$ & ${-}10.8$ & ${-}21.2$ & ${-}5.4$ & ${-}10.6$ & ${-}14.6$ & ${-}6.4$ & ${-}11.0$ & ${-}3.9$ & ${-}10.2$ \\
Qwen3.5 & 2B & ${-}9.4$ & ${-}11.6$ & ${-}7.2$ & ${-}9.5$ & ${-}12.2$ & ${-}23.2$ & ${-}6.6$ & ${-}11.8$ & ${-}16.6$ & ${-}7.3$ & ${-}13.6$ & ${-}4.8$ & ${-}11.2$ \\
Qwen3.5 & 0.8B & ${-}8.7$ & ${-}10.0$ & ${-}6.0$ & ${-}8.6$ & ${-}12.5$ & ${-}20.3$ & ${-}5.7$ & ${-}10.9$ & ${-}14.9$ & ${-}8.2$ & ${-}11.3$ & ${-}5.8$ & ${-}10.3$ \\
Qwen3-VL & 8B & ${-}8.7$ & ${-}8.8$ & ${-}8.2$ & ${-}7.9$ & ${-}10.8$ & ${-}20.9$ & ${-}6.4$ & ${-}8.7$ & ${-}14.1$ & ${-}5.7$ & ${-}10.6$ & ${-}3.4$ & ${-}9.5$ \\
Qwen3-VL & 4B & ${-}12.0$ & ${-}8.9$ & ${-}9.4$ & ${-}10.4$ & ${-}11.2$ & ${-}19.4$ & ${-}5.1$ & ${-}9.3$ & ${-}13.6$ & ${-}6.7$ & ${-}11.3$ & ${-}4.1$ & ${-}10.1$ \\
Cosmos-Reason2 & 32B & ${-}9.9$ & ${-}9.7$ & ${-}8.2$ & ${-}9.6$ & ${-}12.2$ & ${-}19.4$ & ${-}6.3$ & ${-}10.4$ & ${-}13.7$ & ${-}5.6$ & ${-}10.2$ & ${-}4.5$ & ${-}10.0$ \\
Cosmos-Reason2 & 8B & ${-}9.3$ & ${-}9.3$ & ${-}6.1$ & ${-}7.5$ & ${-}10.9$ & ${-}19.9$ & ${-}5.6$ & ${-}10.7$ & ${-}14.4$ & ${-}6.1$ & ${-}11.7$ & ${-}4.9$ & ${-}9.7$ \\
Gemma 4 & 31B & ${-}7.2$ & ${-}8.6$ & ${-}5.2$ & ${-}5.9$ & ${-}9.0$ & ${-}15.5$ & ${-}4.3$ & ${-}8.4$ & ${-}11.1$ & ${-}7.1$ & ${-}9.7$ & ${-}4.6$ & ${-}8.0$ \\
InternVL3.5 & 38B & ${-}6.1$ & ${-}6.7$ & ${-}5.1$ & ${-}5.0$ & ${-}7.7$ & ${-}20.6$ & ${-}4.3$ & ${-}8.1$ & ${-}13.6$ & ${-}7.8$ & ${-}10.8$ & ${-}3.8$ & ${-}8.3$ \\
InternVL3.5 & 14B & ${-}8.9$ & ${-}9.7$ & ${-}6.5$ & ${-}8.2$ & ${-}11.2$ & ${-}22.4$ & ${-}6.7$ & ${-}9.2$ & ${-}14.3$ & ${-}9.9$ & ${-}12.3$ & ${-}4.0$ & ${-}10.3$ \\
InternVL3.5 & 8B & ${-}10.0$ & ${-}9.0$ & ${-}6.6$ & ${-}9.9$ & ${-}11.1$ & ${-}22.9$ & ${-}6.4$ & ${-}10.1$ & ${-}16.4$ & ${-}8.1$ & ${-}11.9$ & ${-}3.9$ & ${-}10.5$ \\
InternVL3.5 & 4B & ${-}10.0$ & ${-}11.3$ & ${-}5.8$ & ${-}8.8$ & ${-}10.9$ & ${-}23.5$ & ${-}6.2$ & ${-}10.4$ & ${-}16.5$ & ${-}7.7$ & ${-}12.7$ & ${-}5.4$ & ${-}10.8$ \\
GLM-4.1V-Thinking & 9B & ${-}9.8$ & ${-}11.9$ & ${-}8.6$ & ${-}12.6$ & ${-}11.9$ & ${-}21.7$ & ${-}6.4$ & ${-}14.8$ & ${-}13.0$ & ${-}7.7$ & ${-}11.1$ & ${-}4.3$ & ${-}11.1$ \\
MiniCPM-V 4.5 & 8B & ${-}7.4$ & ${-}8.8$ & ${-}5.0$ & ${-}8.5$ & ${-}9.5$ & ${-}19.6$ & ${-}5.7$ & ${-}10.0$ & ${-}14.1$ & ${-}8.8$ & ${-}12.3$ & ${-}4.2$ & ${-}9.5$ \\
Ovis2.5 & 9B & ${-}9.6$ & ${-}7.4$ & ${-}4.5$ & ${-}9.2$ & ${-}11.3$ & ${-}21.8$ & ${-}5.0$ & ${-}6.5$ & ${-}15.0$ & ${-}8.2$ & ${-}11.0$ & ${-}4.5$ & ${-}9.5$ \\
Ovis2.5 & 2B & ${-}7.2$ & ${-}9.4$ & ${-}4.6$ & ${-}8.9$ & ${-}11.1$ & ${-}20.8$ & ${-}6.3$ & ${-}8.6$ & ${-}19.4$ & ${-}9.9$ & ${-}12.2$ & ${-}4.1$ & ${-}10.2$ \\
VideoLLaMA3 & 7B & ${-}8.6$ & ${-}8.7$ & ${-}6.3$ & ${-}8.3$ & ${-}12.7$ & ${-}24.2$ & ${-}7.0$ & ${-}12.4$ & ${-}15.9$ & ${-}8.8$ & ${-}14.7$ & ${-}6.4$ & ${-}11.2$ \\
VITA-1.5 & 7B & ${-}7.5$ & ${-}6.4$ & ${-}4.5$ & ${-}8.4$ & ${-}10.9$ & ${-}18.5$ & ${-}5.8$ & ${-}9.6$ & ${-}13.1$ & ${-}8.6$ & ${-}12.1$ & ${-}4.5$ & ${-}9.2$ \\
\midrule
\multicolumn{2}{l}{Mean} & ${-}8.9$ & ${-}9.5$ & ${-}6.4$ & ${-}8.7$ & ${-}10.6$ & ${-}20.1$ & ${-}5.7$ & ${-}9.6$ & ${-}14.1$ & ${-}7.1$ & ${-}11.0$ & ${-}4.4$ & ${-}9.7$ \\
\bottomrule
\end{tabular}}
\end{table*}

\vspace{3pt} \noindent
\textbf{Glass occlusion and defocus cause the largest losses.}
Paired clean-to-corrupted changes vary sharply across corruption types.
Averaged across the 23 models, glass occlusion causes a 20.1-point loss and defocus a 14.1-point loss, compared with 4.4 points for bit errors and 5.7 for overexposure.
Glass occlusion causes the largest drop for all 23 models; defocus ranks second for 22.
The agreement across models shows that these two corruptions are consistently challenging in the evaluated subsets, rather than being driven by a single model's failure.

\vspace{3pt} \noindent
\textbf{Snow causes the largest mean loss among driving corruptions.}
Within the driving set, the mean loss is 9.5 points for snow, followed by 8.9 for rain, 8.7 for fog, and 6.4 for nighttime.
These are paired losses relative to each subset's own clean-video baseline.
However, the corruption types are assigned to different source videos and use fixed presets (Appendix~\ref{app:synthetic-corruptions}), so their ordering describes the evaluated subsets and settings rather than a content-independent difficulty ranking.

\vspace{3pt} \noindent
\textbf{Basic judgments often show larger absolute accuracy drops.}
Across levels, the model-averaged drop is largest at L1
for 11 of the 12 corruptions, with motion blur the exception,
and smallest at L3 for all twelve.
Under glass occlusion, L1 loses 27.1 percentage points,
compared with 14.0 at L3.
These results show that lower reasoning demands do not
necessarily imply smaller absolute losses under degradation.
However, the magnitude of these drops also depends on
clean accuracy: lower clean accuracy at L3 may leave less
room for further decline.
The observed differences therefore do not by themselves
establish greater intrinsic corruption sensitivity at L1.
These per-corruption statistics average models within each
corruption; the overall $\Delta$ in
Table~\ref{tab:grape_main_levels} instead weights questions
across corruptions, whose subset sizes differ.

\subsection{Accuracy across Difficulty Levels}
\label{app:level-profiles}

\noindent
\textbf{Empirical accuracy need not follow a strict level ordering.}
On the real-world set, 23 of the 24 open-weight configurations exhibit $\mathrm{L1}>\mathrm{L2}>\mathrm{L3}$ in aggregate accuracy (Table~\ref{tab:grape_realworld_levels}).
Kimi K3 is the exception, scoring higher at L2 than at L1 ($88.8\%$ vs.\ $86.7\%$), although L3 remains the lowest-scoring level for all 24 configurations.
Local reversals also occur: under real fog, 11 of these configurations score higher at L3 than at L2.
For example, Qwen3.5-27B achieves $92.0\%$ at L3 compared with $88.0\%$ at L2.
The levels describe intended reasoning demands without imposing a universal empirical accuracy ordering.
These comparisons alone cannot explain whether reversals arise from model behavior, the question mix, or annotation difficulty; resolving this requires auditing individual questions.

\subsection{Reliability of Safety-Relevance Ratings}
\label{app:safety-rater}

The safety-relevance ratings in Section~\ref{sec:motivation} are produced by Claude Opus~5.5 from the question text and answer options alone.
To test whether they depend on the choice of rater, we re-rate all 1,716 real-world questions with GPT-6 Sol using the identical rubric, inputs, and question order.

\begin{figure}[t]
    \centering
    \includegraphics[width=\columnwidth]{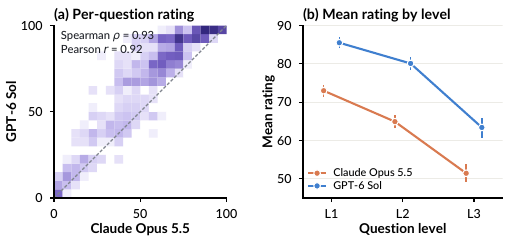}
    \caption{\textbf{Agreement between two safety-relevance raters.} (a) Per-question ratings of the 1,716 real-world questions by Claude Opus~5.5 and GPT-6 Sol (darker cells contain more questions; dashed line: identity). (b) Mean rating per question level with 95\% bootstrap intervals.}
    \label{fig:safety-rater}
\end{figure}

\vspace{3pt} \noindent
\textbf{The two raters order questions consistently.}
The ratings are strongly correlated (Spearman $\rho=0.93$, Pearson $r=0.92$; Figure~\ref{fig:safety-rater}a), and the agreement holds within each level ($\rho=0.93$, $0.91$, and $0.95$ for L1, L2, and L3).
GPT-6 Sol assigns higher absolute ratings on average (76.3 vs.\ 63.1), so the four rating bins agree less exactly (quadratic-weighted $\kappa=0.69$).

\vspace{3pt} \noindent
\textbf{Both raters show the same level trend.}
Under both raters, the mean rating decreases from L1 to L3 (Opus: 73.0, 64.9, 51.5; GPT-6 Sol: 85.5, 80.1, 63.4; Figure~\ref{fig:safety-rater}b).
Within question groups, the L1 question is rated higher than the L3 question in 408 of 572 groups by Opus and in 402 by GPT-6 Sol (two-sided sign tests, $p<10^{-27}$).

\subsection{Proprietary and Open-Weight Comparisons}
\label{app:proprietary-comparisons}

Table~\ref{tab:grape_realworld_levels} also supports comparisons between the six proprietary configurations and the open-weight models.

\vspace{3pt} \noindent
\textbf{Proprietary gains vary by condition and level.}
GPT-5.6 Sol leads overall ($87.8\%$), but its advantage over
the strongest open-weight model, Kimi K3 ($85.4\%$), varies
from 1.0 point in fog to 3.7 points in snow.
The advantage also differs across question levels: 3.5 points
at L1, 1.0 at L2, and 2.6 at L3.
Its overall lead therefore combines uneven gains across
conditions and reasoning demands.
All six proprietary configurations score lower at L3 than
at both L1 and L2; even GPT-5.6 Sol reaches $83.4\%$ at L3,
compared with $90.2\%$ at L1.
Moreover, Kimi K3 exceeds GPT-6 Luna and Claude Sonnet 5.5
overall, showing that proprietary status alone does not
predict superiority.

\section{Synthetic-to-Real Training and Analysis}
\label{sec:supp_training}

\subsection{Data Splits and Adaptation}

\noindent
\textbf{Data and splits.}
For the transfer experiments in Section~\ref{sec:exp},
all training uses the synthetic set.
We split its 10,841 chains by source video and hold out
5\% (533 chains, 1,599 questions).
The remaining 10,308 chains (30,924 questions) form the
training partition; each L1, L2, and L3 question of a
chain is one training sample.
The held-out chains are divided into a validation split
(299 chains, 897 questions), used for hyperparameter
and variant selection, and a test split
(234 chains, 702 questions), used only for reporting.
The training, validation, and test splits are mutually
disjoint by source video.
The main transfer comparisons use the final checkpoint
of each training run.
For these transfer experiments, the real-world set
(572 videos, 1,716 questions) is used exclusively for
testing: no real-world video, question, or label enters
training, model selection, or hyperparameter tuning.

\vspace{3pt} \noindent
\textbf{Backbone and adaptation.}
All variants start from Qwen3.5-9B and are adapted with LoRA (rank 16, $\alpha{=}32$, dropout 0.05) on the attention and MLP projections of the language model, including the linear-attention projections; the vision encoder is frozen.
Each training video is represented by 32 frames sampled at uniform bin centers and resized to fit within $640{\times}360$ pixels (sides multiple of 32).
The corrupted and clean frame sets of a chain use the same frame indices, so student and teacher inputs differ only in pixels.
Losses are applied only to response tokens.

\subsection{Training Recipes and Distillation}
\label{app:transfer-objectives}

The training objectives are given in Equations~\eqref{eq:sft}--\eqref{eq:opsd}.

\vspace{3pt} \noindent
\textbf{SFT and reasoning SFT.}
We use AdamW (no weight decay), a peak learning rate of $10^{-4}$ with 3\% linear warm-up and cosine decay, gradient clipping at 1.0, and an effective batch size of 16, for one epoch in bfloat16 on four H100 GPUs.
SFT predicts the option letter only.
For reasoning SFT, rationales are generated by the unadapted model on the clean video with a prompt asking for at most two sentences of reasoning followed by ``Answer: X''; a rationale is accepted if its final answer is correct.

\vspace{3pt} \noindent
\textbf{On-policy self-distillation.}
The student $\pi_\theta$ and the frozen teacher $\pi_T$ are both initialized from the reasoning-SFT model. Let $x=(\tilde{v},q,\mathcal{A})$ denote the corrupted input, $x^0=(v^0,q,\mathcal{A})$ its clean counterpart, and $y$ the reference answer. Each step samples one response $\hat{z}\sim\pi_\theta(\cdot\mid x)$ per question from the current student (temperature 1.0, at most 96 new tokens). For the reported variant, the teacher receives both $x^0$ and $y$. Teacher and student score the same sampled prefixes, with logits aligned token by token, giving
\begin{equation}
D_t=\mathrm{KL}\big(\pi_\theta(\cdot\mid x,\hat{z}_{<t})\,\big\|\,\pi_T(\cdot\mid x^0,y,\hat{z}_{<t})\big).
\label{eq:opsd_kl}
\end{equation}
Reverse KL is computed over the full vocabulary and averaged over response tokens. The answer term is
\begin{equation}
\ell_{\mathrm{ans}}=-\log\pi_\theta(y\mid x,\hat{z}_{<a}),
\end{equation}
where $a$ is the final answer position in $\hat{z}$. The expectation in Equation~\eqref{eq:opsd} averages over training questions and on-policy responses.

We use reverse KL with weight 1.0, an answer cross-entropy weight $\lambda=0.5$, a learning rate of $5{\times}10^{-5}$, and one pass over the 30,924 training questions. We compared two teachers, one that sees only the clean video and one that additionally receives the reference answer in its prompt; the latter has higher synthetic validation accuracy (87.5\% vs.\ 87.3\%) and is reported in the main paper (real-world: 80.4\% vs.\ 80.8\%). OPSD costs about 17.6 H100 GPU-hours, roughly twice the cost of SFT.

\subsection{Transfer Evaluation and Analysis}
\label{sec:exp_scaling}

\noindent
\textbf{Evaluation protocol.}
At evaluation time, the base and fine-tuned models use
the Qwen3.5-9B video-input protocol described in
Appendix~\ref{app:evaluation-protocol}, including the
question-specific cutoffs on the real-world set.
The 32-frame inputs described above are used during training.
Fine-tuned LoRA adapters are merged into the model weights
before evaluation.
Inference uses vLLM 0.31.0 in BF16, with greedy decoding
and thinking disabled.
The base model and direct SFT use an answer-only prompt
with a maximum output length of 1,024 tokens.
Reasoning SFT and OPSD use the reasoning prompt described
above, requesting at most two short reasoning sentences
followed by \texttt{Answer: X}, with a maximum output
length of 512 tokens.
For these reasoning variants, the final answer is extracted
from the last \texttt{Answer: X} match, falling back to
the last standalone valid option letter.
The reported direct-SFT results are averaged over three
training runs.
Reasoning SFT and OPSD are each evaluated from a single
training run.

\vspace{3pt} \noindent
\textbf{Transfer across conditions and methods.}
Direct SFT improves the real-world accuracy of Qwen3.5-9B
from 76.7\% to 81.1\%.
The gains are 5.5 percentage points on rain, 4.1 on snow,
4.5 on night, and 2.8 on fog.
After direct SFT, Qwen3.5-9B achieves higher real-world accuracy
than the Qwen3.6-27B and Cosmos-Reason2-32B baselines.
Its accuracy is also close to that of Qwen3.5-27B,
as reported in Table~\ref{tab:grape_realworld_levels}.
Reasoning SFT raises real-world accuracy to 78.1\%.
OPSD improves overall accuracy by 2.2 percentage points
over reasoning SFT.
Its L1 accuracy reaches 86.9\%.
OPSD's overall accuracy remains below that of direct SFT.

\vspace{3pt} \noindent
\textbf{Data scaling.}
Figure~\ref{fig:synthetic-to-real-transfer}(a) uses level-balanced training subsets at learning rate $10^{-4}$ and reports per-level real-world accuracy averaged over three seeds, with $\pm1$ standard deviation. The horizontal axis counts training-question exposures: one full pass contains 30,924 samples, and the second pass revisits the same pool.
L1 and L2 obtain most of their gains within the first 3K training questions, while larger subsets provide additional gains for L3. Extending training from one to two passes lowers the mean accuracy at all three levels.

\vspace{3pt} \noindent
\textbf{Learning-rate and epoch sweeps.}
Figure~\ref{fig:synthetic-to-real-transfer}(b) compares final SFT models across learning rates $\{10^{-5}, 3{\times}10^{-5}, 10^{-4}, 3{\times}10^{-4}\}$ and $\{0.5, 1, 2\}$ epochs. Each line fixes the epoch budget and varies the learning rate; the two axes report synthetic validation and real-world accuracy.

\subsection{Steering Details}
\label{app:steering-details}

\noindent\textbf{Model and intervention.}
Both panels of Figure~\ref{fig:perception-reasoning-steering} start from the original Qwen3.5-9B checkpoint, before any fine-tuning on \grapebench{}.
We freeze the backbone and initialize the two 4,096-dimensional vectors $\mathbf{u}$ and $\mathbf{d}$ to zero, giving 8,192 trainable parameters per domain.
At the output of decoder layer 12 (zero-indexed, out of 32), we add $\mathbf{s}(\lambda)=\mathbf{u}+\lambda\mathbf{d}$ only to text positions after the final video token, covering the question, answer options, and answer prompt.

\vspace{3pt} \noindent
\textbf{Synthetic training questions.}
For panel~(a), we start from 2,871 VRU-Accident question chains in the synthetic training split, covering rain, snow, night, and fog.
Each input uses the same 32 pre-extracted $640\times352$ frames as the SFT experiment.
We estimate the difficulty of each question from the fraction of 22 other models in Table~\ref{tab:grape_main_levels} that answer it correctly, excluding Qwen3.5-9B.
We trim the lower tail of L1 accuracy and the upper tail of L3 accuracy using their respective decile cutoffs.
This removes L1 questions answered correctly by fewer than five of the 22 models and L3 questions answered correctly by all 22, leaving 2,644 L1 and 2,743 L3 questions.
We randomly sample 1,000 questions per level from these pools. This filter applies only to steering training.

\vspace{3pt} \noindent
\textbf{Preference pairs and updates.}
Before training, we cache the unsteered model's next-token log probabilities for the four option letters on each selected question.
We pair the correct letter with the highest-scoring incorrect letter and keep the pair fixed throughout training.
Reference accuracy on the two synthetic training pools is 86.0\% for L1 and 64.3\% for L3.
At each optimizer step, we draw one $\lambda\sim\mathcal{U}[-1,1]$ and share it across two L1 and two L3 questions, accumulating gradients one question at a time.
We apply the preference loss in Section~\ref{sec:method} with $\beta=0.1$. The weights $(1+\lambda)/2$ and $(1-\lambda)/2$ set the relative contributions of the L1 and L3 losses, respectively. Both reward a larger margin for the correct letter.
We train on one GPU with gradient checkpointing and AdamW (peak learning rate $5\times10^{-4}$, weight decay 0.05), using 25 warmup steps followed by cosine decay.
Three passes over the 2,000 synthetic questions give 1,500 optimizer steps.
The final vectors have norms $\|\mathbf{u}\|=2.77$ and $\|\mathbf{d}\|=1.13$, with cosine similarity 0.19.
The L1 preference margin peaks at 21.6 for positive $\lambda$, while the L3 margin ranges from 8.2 to 8.3 across settings.

\vspace{3pt} \noindent
\textbf{Held-out evaluation.}
The synthetic evaluation uses 129 driving question chains from the validation and test splits, disjoint by video from training. Each chain contributes one question per level, giving 387 questions including L2, which is held out of steering training.
We score every question under ten settings: the unsteered baseline and nine values of $\lambda$ from $-1$ to $+1$ in increments of 0.25.
At $\lambda=0$, the shared vector $\mathbf{u}$ remains active.
For each setting, we predict the most probable of the four next-token option letters and compute overall accuracy over the pooled L1, L2, and L3 predictions.
For panel~(b), we sample 300 L1 and 300 L3 questions from the real-world driving training split (600 questions total), train a separate vector pair using the same protocol, and evaluate on held-out videos.

\section{Extended Related Work and Our Unique Contributions}
\label{app:extended-related}

\subsection{Video Understanding and Reasoning}

\noindent\textbf{Evaluation benchmarks.}
Video benchmarks span broad capability assessment and detailed reasoning evaluation~\cite{li2024mvbench,fu2025videomme,nagrani2025minerva}. NExT-QA and STAR emphasize temporal, causal, and situated reasoning~\cite{xiao2021nextqa,wu2021star}; MVBench, Video-MME, and EgoSchema broaden task, domain, and temporal coverage~\cite{li2024mvbench,fu2025videomme,mangalam2023egoschema}. VCR-Bench and MINERVA include reasoning traces~\cite{qi2025vcrbench,nagrani2025minerva}, while VideoReasonBench and Video-MME-v2 further structure reasoning and capability evaluation~\cite{liu2026videoreasonbench,fu2026videommev2}. Sensitivity studies also examine how video-language models answer when temporal or visual evidence is altered~\cite{gou2026howvideo,tv2026reveal}. \grapebench{} examines how degradation affects performance across question levels and how intended difficulty relates to safety relevance.

\vspace{3pt} \noindent
\textbf{Video reasoning models.}
LLaVA-Video expands video instruction tuning through generated captioning and QA data~\cite{zhang2025llavavideo}. Qwen3-VL advances long-context understanding and temporal alignment~\cite{bai2025qwen3vl}, while InternVL3.5 improves multimodal reasoning through reinforcement learning~\cite{wang2025internvl35}. Apollo studies design choices for video input and model architectures~\cite{zohar2025apollo}. Video-R1 and VideoChat-R1 reinforce temporal reasoning and spatiotemporal perception~\cite{feng2025videor1,li2025videochatr1}, while VideoChat-R1.5 increases perceptual computation through iterative inspection of spatiotemporal regions~\cite{yan2025videochatr15}. Such advances motivate testing whether stronger reasoning remains reliable when visual evidence degrades. Our condition- and level-specific evaluations reveal whether aggregate improvements extend to fundamental, safety-relevant judgments.

\subsection{Robustness and Adverse Driving Conditions}

\noindent\textbf{Controlled visual degradations.}
ImageNet-C systematized common-corruption evaluation~\cite{hendrycks2019imagenetc}, with subsequent work extending robustness analysis to spatiotemporal and video-language models~\cite{yi2021minikineticsc,schiappa2023kinetics400p,schiappa2022msrvttp}. R-Bench and MLLM-IC examine image distortions and fine-grained multimodal capabilities~\cite{li2025rbench,qiu2025mllmic}; MVTamperBench and PVRBench evaluate temporal tampering and perturbed video reasoning~\cite{agarwal2025mvtamperbench,he2026rova}. Robust-R1, Robust-U1, and ROVA explore degradation-aware reasoning, visual recovery, and clean-to-corrupted consistency~\cite{tang2026robustr1,tang2026robustu1,he2026rova}. \grapebench{} combines paired clean-to-corrupted comparisons with natural-video evaluation, distinguishing controlled corruption losses from performance under real-world conditions.

\vspace{3pt} \noindent
\textbf{Naturally adverse driving conditions.}
ACDC, Boreas, and Seeing Through Fog provide observations under adverse weather or across seasons~\cite{sakaridis2021acdc,burnett2023boreas,bijelic2020fog}. SHIFT supplies controlled synthetic domain shifts~\cite{sun2022shift}, and driving perception benchmarks isolate corruptions in LiDAR, camera-based bird's-eye-view inputs, or multiple sensor streams~\cite{kong2023robo3d,xie2025robobev,hao2025mscbench}. DriveLM and DriveVLM connect visual reasoning with perception or planning tasks~\cite{sima2024drivelm,tian2025drivevlm}, and DriveBench tests reliability under clean, corrupted, and text-only inputs~\cite{xie2025drivebench}. ObsDriveBench studies observability, spatial reliability, and risk-aware decisions using camera, LiDAR, and radar observations in adverse weather~\cite{yan2026obsdrivebench}. Our real-world partition focuses on video QA using the available video prefix. Its three-level design and auxiliary safety-relevance ratings support analysis of both reasoning demands and the significance of the judgments affected.

\subsection{Adaptation and Preference Control}

\noindent\textbf{Synthetic supervision and distillation.}
Generated instruction data~\cite{zhang2025llavavideo} and filtered rationales~\cite{zelikman2022star} support multimodal and reasoning supervision. Earlier approaches use teacher predictions for distillation, train students on augmented inputs with teacher labels, or encourage consistency across augmentations~\cite{hinton2015distill,xie2020noisystudent,hendrycks2020augmix}. Our synthetic corruptions provide a clean counterpart to each degraded input for teacher supervision. Imitation learning addresses distribution mismatch by collecting feedback on states visited by the learner~\cite{ross2011dagger}. Similarly, on-policy distillation uses teacher feedback on student-generated responses~\cite{agarwal2024onpolicy}. We adapt this principle by having a frozen clean-video teacher score the student's corrupted-input reasoning trajectories. Comparing this recipe with direct and reasoning-based SFT tests whether the synthetic partition supports transfer to naturally degraded videos.

\vspace{3pt} \noindent
\textbf{Preference-conditioned steering.}
DPO optimizes preference pairs~\cite{rafailov2023dpo}; MODPO and Panacea address multiple objectives and varying preferences~\cite{zhou2024modpo,zhong2024panacea}; and BiPO learns representation-level steering vectors~\cite{cao2024bipo}. Other methods control preferences with attribute conditioning or interpolate models optimized for different rewards~\cite{dong2023steerlm,rame2023rewardedsoups}. We draw on these ideas to control an accuracy trade-off exposed by the benchmark. Two learned vectors define a shared intervention and a preference-dependent direction in a frozen VLM, shifting emphasis between basic judgments and demanding reasoning. Separate synthetic and real-world experiments evaluate held-out control, reporting overall accuracy alongside level-specific changes.

\subsection{Our Unique Contributions}

\noindent\textbf{Which judgments hold up when vision degrades?}
PVRBench tests reasoning under controlled video perturbations~\cite{he2026rova}, ObsDriveBench studies perception and risk under naturally adverse driving conditions~\cite{yan2026obsdrivebench}, and VideoReasonBench probes different levels of reasoning~\cite{liu2026videoreasonbench}.
\grapebench{} brings these questions into one evaluation. Each video has three independently answerable questions of increasing reasoning demand, both on the synthetic set spanning twelve corruptions across daily-life and driving videos and on driving footage recorded in rain, snow, fog, or at night.
We then used Claude Opus~5.5 to rate the safety relevance of the completed real-world questions. L1 questions received the highest ratings on average. For example, recognizing whether a car stopped for a red light can matter more to a driving decision than counting the vehicles queued beside it.

\vspace{3pt} \noindent
\textbf{What does an overall score conceal?}
We evaluate 23 open-weight models across both sets and additional proprietary frontier models on real-world footage. Models can reverse rank between simulated rain and real rain. Rankings also depend on which judgments are scored. On real-world videos, Kimi K3 leads Qwen3.8-Flash-Next overall but trails it on L1.
Larger models generally perform better, although scaling alone does not guarantee higher real-world accuracy or smaller losses under corruption.

\vspace{3pt} \noindent
\textbf{Can training improve or steer these judgments?}
Paired clean and corrupted videos let us test whether synthetic supervision transfers to real driving footage. Direct SFT raises Qwen3.5-9B's real-world accuracy from 76.7\% to 81.1\%, although a better fit to synthetic data can hurt transfer.
Building on on-policy distillation~\cite{agarwal2024onpolicy}, we give the teacher the clean video and have it score reasoning steps produced by the student on the corrupted video. This raises real-world accuracy from 78.1\% with reasoning SFT to 80.4\%.
Building on representation steering~\cite{cao2024bipo}, we also learn two vectors for the frozen Qwen3.5-9B and vary a single coefficient to shift accuracy between basic, safety-relevant judgments and more demanding reasoning. We observe this trade-off on held-out videos in both domains, using vectors trained separately in each, with little change in overall accuracy.

\section{Limitations and Societal Impact}
\label{app:limitations-impact}

\subsection{Limitations}

\noindent\textbf{Corruption severity.}
We evaluate each synthetic corruption at a single intensity.
We will release code for adjusting the severity of each corruption so that researchers can render the same clips at several intensities.
Holding the clips and questions fixed would make it possible to measure how rapidly accuracy declines with severity at each question level.

\vspace{3pt} \noindent
\textbf{Inference efficiency.}
Our evaluation focuses on answer accuracy under visual degradation.
Driving systems must also make decisions in real time, so the time a model takes to answer matters alongside its accuracy.
Measuring end-to-end latency and peak memory on the same hardware would help identify open-weight models that maintain accuracy within a given response time and memory budget.

\subsection{Societal Impact}

\noindent\textbf{Potential benefits and deployment limits.}
By exposing failures under degraded visual evidence and separating performance across difficulty levels, \grapebench{} may help researchers identify weaknesses that aggregate scores conceal. Synthetic training and lightweight steering provide tools for studying how to address these weaknesses. However, improved benchmark accuracy or a preference for L1 questions does not establish safer driving behavior. Deployment decisions require evaluation of actual task outcomes, uncertainty, and failure handling under relevant operating conditions; the benchmark is a research instrument rather than a safety certification.

\vspace{3pt} \noindent
\textbf{Privacy, representation, and misuse.}
Source videos can contain identifiable people, vehicles, and locations. Reuse and redistribution should respect source-data terms and minimize unnecessary exposure of personal information. Uneven coverage of regions, conditions, and road users may also hide failures affecting underrepresented groups, motivating broader data collection and subgroup evaluation. More robust video understanding could support intrusive surveillance, while steering could be used to improve selected metrics at the expense of unreported capabilities. Transparent reporting across conditions and difficulty levels is therefore important when interpreting or reusing these methods.

\end{document}